\documentclass[nohyperref]{article}

\usepackage{microtype}
\usepackage{graphicx}
\usepackage{subfigure}
\usepackage{booktabs} % for professional tables

\usepackage[numbers]{natbib}
\usepackage{hyperref}
\usepackage{url}
\usepackage{arydshln}
\usepackage[acronyms]{glossaries}
\usepackage{caption}
\usepackage{amsmath}
\usepackage{amssymb}
\usepackage{color,soul}
\usepackage[capitalize,noabbrev]{cleveref}
\usepackage{tikz}
\usepackage{array}
\newcolumntype{K}[1]{>{\centering\arraybackslash}p{#1}}
\usepackage{enumerate} 

\usepackage[acronyms]{glossaries}
\usepackage{wrapfig,lipsum}
\usetikzlibrary{shapes,arrows,arrows.meta}

\definecolor{greenish}{RGB}{23, 127, 117}
\definecolor{yellowish}{RGB}{182, 119, 33}
\definecolor{redish}{RGB}{182, 33, 45}
\definecolor{lightgreenish}{RGB}{33, 182, 168 }

\Crefname{equation}{Eqn.}{Eqns.}
\Crefname{figure}{Fig.}{Figs.}
\Crefname{table}{Tab.}{Tabs.}
\Crefname{section}{Sec.}{Sections}

\usepackage{glossaries}

\newacronym{node}{neural ODE}{neural ordinary differential equation}
\newacronym{mlp}{MLP}{multi-layer perceptron}

\newacronym{rkn}{RKN}{recurrent Kalman network}
\newacronym{cru}{CRU}{continuous recurrent unit}
\newacronym{fcru}{f-CRU}{fast CRU}
\newacronym{gru}{GRU}{gated recurrent unit}
\newacronym{lstm}{LSTM}{long short-term memory network}
\newacronym{sde}{SDE}{stochastic differential equation}
\newacronym{rnn}{RNN}{recurrent neural network}
\newacronym{cdkf}{cdKF}{continuous-discrete Kalman filter}
\newacronym{vae}{VAE}{variational autoencoder}
\newacronym{mse}{MSE}{mean squared error}
\newacronym{nll}{NLL}{negative loglikelihood}
\newacronym{icu}{ICU}{intensive care unit}
\newacronym{ode}{ODE}{ordinary differential equation}
\newacronym{ushcn}{USHCN}{United States Historical Climatology Network}
\newacronym{nsde}{neural SDE}{neural SDE}
\newacronym{mtand}{mTAND}{multi-time attention network}

\glsdisablehyper

\usepackage{hyperref}
\usepackage{enumitem}

\usepackage[accepted]{icml2022}

\icmltitlerunning{Modeling Irregular Time Series with Continuous Recurrent Units}
\begin{document}

\twocolumn[
\icmltitle{Modeling Irregular Time Series with Continuous Recurrent Units}




\icmlsetsymbol{intern}{*}

\begin{icmlauthorlist}
\icmlauthor{Mona Schirmer}{MS}
\icmlauthor{Mazin Eltayeb}{ME}
\icmlauthor{Stefan Lessmann}{SL}
\icmlauthor{Maja Rudolph}{MR}
\end{icmlauthorlist}

%\icmlaffiliation{HU}{Humboldt-Universität zu Berlin, Germany}
%\icmlaffiliation{BCAI_DE}{Bosch Center for Artificial Intelligence, Germany}
%\icmlaffiliation{BCAI_US}{Bosch Center for Artificial Intelligence, USA}
%\icmlaffiliation{intern}{Work done during an internship at Bosch Center for Artificial Intelligence}
\icmlaffiliation{MS}{Humboldt-Universität zu Berlin, Germany. Work done during an internship at Bosch Center for AI}
\icmlaffiliation{ME}{Bosch Center for AI, Germany}
\icmlaffiliation{SL}{Humboldt-Universität zu Berlin, Germany}
\icmlaffiliation{MR}{Bosch Center for AI, USA}

\icmlcorrespondingauthor{Mona Schirmer}{mona.schirmer@ensae.fr}
\icmlcorrespondingauthor{Maja Rudolph}{maja.rudolph@us.bosch.com}

\icmlkeywords{Machine Learning}

\vskip 0.3in
]



\printAffiliationsAndNotice{}  % leave blank if no need to mention equal contribution
%\printAffiliationsAndNotice{} % otherwise use the standard text.

\begin{abstract}
\Glspl{rnn} are a popular choice for modeling sequential data.  Modern \gls{rnn} architectures assume constant time-intervals between observations. However, in many datasets (e.g. medical records) observation times are irregular and can carry important information. To address this challenge, we propose \glspl{cru} -- a neural architecture that can naturally handle irregular intervals between observations. The \gls{cru} assumes a hidden state, which evolves according to a linear stochastic differential equation and is integrated into an encoder-decoder framework. The recursive computations of the \gls{cru} can be derived using the continuous-discrete Kalman filter and are in closed form. The resulting recurrent architecture has temporal continuity between hidden states and a gating mechanism that can optimally integrate noisy observations. We derive an efficient parameterization scheme for the CRU that leads to a fast implementation \acrshort{fcru}. We empirically study the \gls{cru} on a number of challenging datasets and find that it can interpolate irregular time series better than methods based on neural ordinary differential equations. 
%% OLD/NEW HYBRID ABSTRACT
%\Glspl{rnn} are a popular choice for modeling sequential data. Standard \glspl{rnn} assume constant time-intervals between observations. However, in many applications such as health care, observation times are irregular and can carry important information. To address this challenge, we propose \glspl{cru} - a neural architecture that can naturally handle irregular intervals between observations. The \gls{cru} assumes a hidden state, which evolves according to a linear stochastic differential equation and is integrated into an encoder-decoder framework. The recursive computations of the \gls{cru} can be derived using the continuous-discrete Kalman filter and are in closed form. The resulting recurrent architecture has temporal continuity between hidden states and a gating mechanism that can optimally integrate noisy observations. We derive an efficient parameterization scheme for the \gls{cru} that leads to a fast implementation. In an empirical study, we show that the \gls{cru} can better interpolate irregular time series than \gls{node}-based models. We also show that our model can infer dynamics from images and that the Kalman gain singles out valuable candidates for state updates from noisy observations.
\end{abstract}

\section{Introduction}
\glsresetall
Recurrent architectures, such as the \gls{lstm} \citep{hochreiter1997long} or \gls{gru} \citep{chung2014empirical} have become a principal machine learning tool for modeling time series. Their modeling power comes from a hidden state, which is recursively updated to integrate new observations, and a gating mechanism to balance the importance of new information with history already encoded in the latent state. 

Although continuous formulations were frequently considered in early work on \glspl{rnn} \citep{pineda1987generalization, pearlmutter1989learning, pearlmutter1995gradient}, modern \glspl{rnn} typically assume regular sampling rates \citep{hochreiter1997long,chung2014empirical}. Many real world data sets, such as electronic health records or climate data, are irregularly sampled. Measurements of a patient's health status, for example, are only available when the patient sees a doctor. Hence, the time between observations also carries information about the underlying time series. A lab test not administered for many months could imply that the patient was doing well in the meantime, while frequent visits might indicate that the patient's health is deteriorating. Discrete \glspl{rnn} face difficulties modeling such data as they do not reflect the continuity of the underlying temporal processes. 

Recently, the work on \glspl{node} \citep{chen2018neural} has established an elegant and practical way of modeling irregularly sampled time series. Recurrent architectures based on \glspl{node} determine the hidden state between observations by an \gls{ode} and update its hidden state at observation times using standard \gls{rnn} gating mechanisms (\citealt{rubanova2019latent, brouwer2019gru, lechner2020learning}). These methods typically rely on some form of numerical \gls{ode}-solver, a network component that can prolong training time significantly (\citealt{rubanova2019latent,shukla2020survey}).

%In both standard and \gls{ode}-based \glspl{rnn}, the hidden state is designed as a deterministic component as opposed to a random variable. A probabilistic state space, however, benefits from theoretical grounds for optimal solutions and allows tracking uncertainty concisely over the transformation steps of observations. Reliable uncertainty quantification is indispensable in decision making contexts such as autonomous driving or AI-based medicine. 

We propose \gls{cru}, a probabilistic recurrent architecture for modelling irregularly sampled time series. An encoder maps observations into a latent space, which is governed by a linear \gls{sde}. The analytic solution for propagating the latent state between observations and the update equations for integrating new observations are given by the continuous-discrete formulation of the Kalman filter. Employing the linear \gls{sde} state space model and the Kalman filter has three advantages. First, a probabilistic state space provides an explicit notion of uncertainty for an uncertainty-driven gating mechanism and for confidence evaluation of predictions. Second, as the Kalman filter is the optimal solution for the linear filtering problem \citep{kalman1960new}, the gating mechanism is optimal in a locally linear state space. Third, the latent state at any point in time can be resolved analytically, therefore bypassing the need for numerical integration techniques or variational approximations. In summary, our contributions are as follows:
\vspace{-5pt}
\begin{itemize}[leftmargin=*,itemsep=0pt]
    \item[\textbullet] In \Cref{sec:method}, we develop the \gls{cru}, a model that combines the power of neural networks for feature extraction with the advantages of probabilistic state-space models, specifically the continuous-discrete  Kalman filter. As a result, the \gls{cru} is a powerful neural architecture that can naturally model data with irregular observation times. A PyTorch implementation is available on github.\footnote{\url{https://github.com/boschresearch/Continuous-Recurrent-Units}}
    \item[\textbullet] In \Cref{sec:params}, we derive a novel parameterization of the latent state transition matrices via their eigendecomposition leading to a faster implementation we call \gls{fcru}.% and to trade off speed with accuracy.
    \item In \Cref{sec:exp}, we study the \gls{cru} on electronic health records, climate data and images. We find that 
(i) our method can better interpolate irregular time series than \gls{node}-based methods,
(ii) the \gls{cru} can handle uncertainty arising from both noisy and partially observed inputs,
     (iii) \gls{cru} outperforms both discrete \gls{rnn} counterparts and \gls{node}-based models on image data.
\end{itemize}

\section{Related Work}
\label{sec:related}
\paragraph{Stochastic \glspl{rnn}} \Glspl{rnn}, such as \glspl{lstm} or \glspl{gru}, are powerful sequence models \citep{hochreiter1997long,chung2014empirical}, but due to the lack of stochasticity in their internal transitions, they may fail to capture the variability inherent in certain data 
%of certain modeling problems
(\citealt{chung2015recurrent}). While there are various stochastic \glspl{rnn} \citep[e.g.][]{bayer2014learning, fraccaro2016sequential,goyal2017z,schmidt2018deep}, our work is most closely related to deep probabilistic approaches based on Kalman filters \citep{kalman1960new}.
Variations on deep Kalman filters (\citealt{krishnan2015deep,karl2016deep,fraccaro2017disentangled}) typically require approximate inference, but \citet{becker2019recurrent} employ a locally linear model in a high-dimensional factorized latent state for which the
Kalman updates can be obtained in closed form. By extending this approach with a continuous latent state, the \gls{cru} can model sequences with irregular observation times.

\paragraph{\Glspl{rnn} for Irregular Time Series}
Applying discrete \glspl{rnn} to irregularly sampled time series requires the discretization of the time line into uniform bins. This often reduces the number of observations, may result in a loss of information, and evokes the need for imputation and aggregation strategies. To avoid such preprocessing, \citet{choi2018mime} and \citet{mozer2017discrete} propose to augment observations with timestamps. \citet{lipton2016directly} suggest observation masks. 
However, such approaches have no notion of dynamics between observations. An alternative approach is to decay the hidden state exponentially between observations according to a trainable decay parameter \citep{che2018recurrent,cao2018brits}. These methods are limited to decaying dynamics, %as they enforce that the encoded history becomes less relevant as time passes. T
whereas the \gls{cru} is more expressive. 

\paragraph{Continuous-Time \glspl{rnn}}
%What is today most commonly referred to as a \gls{rnn} has, in the early days, been considered the discrete special case of continuous-time \glspl{rnn} \citep{pearlmutter1995gradient}.
Continuous-time \glspl{rnn} have a long history, dating back to some of the original work on recurrent networks in the field. They are recurrent architectures whose internal units are governed by a system of \glspl{ode} with trainable weights \citep{pearlmutter1995gradient}. The theory for different gradient-based optimization schemes for their parameters have been developed by \citet{pineda1987generalization}, \citet{pearlmutter1989learning}, and \citet{sato1990learning}. Notably, \citet{lecun1988theoretical}'s derivation  using the adjoint method provides the theoretical foundation for modern implementations of \glspl{node} \citep{chen2018neural}.

\paragraph{\Glspl{node}} \Glspl{node} model the  continuous dynamics of a hidden state by an \gls{ode} specified by a neural network layer. \citet{chen2018neural} propose latent \gls{ode}, a generative model whose latent state evolves according to a \gls{node}. However, it has no update mechanism to incorporate incoming observations into the latent trajectory. \citet{kidger2020neural} and \citet{morrill2021neural} extend \glspl{node} with concepts from rough analysis, which allow for online learning. \gls{ode}-\gls{rnn} \citep{rubanova2019latent} and \gls{ode}-\gls{lstm} \citep{lechner2020learning} use standard \gls{rnn} gates to sequentially update the hidden state at observation times. \gls{gru}-\gls{ode}-B \citep{brouwer2019gru} and Neural Jump \gls{ode} \citep{herrera2020neural} couple \gls{ode} dynamics with an Bayesian update step. %As \citet{herrera2020neural} point out, there is no theoretical ground that \gls{gru}-\gls{ode}-B or \gls{ode}-\gls{rnn} yield optimal predictions. \citet{herrera2020neural} prove convergence of their \gls{njode} model to the optimal prediction under Markovian assumptions. 
\Gls{node} approaches typically rely on a numerical \gls{ode} solver, whereas the state evolution of a \gls{cru} is in closed form. 
%In addition, unlike our approach, they typically do not have a principled way to deal with uncertainty or observation noise. 

\paragraph{Neural \glspl{sde}} As the stochastic analogue of \glspl{node}, neural \glspl{sde} define a latent temporal process with a \gls{sde} parameterized by neural networks. \citet{li2020scalable} use the stochastic adjoint sensitive method to compute efficient gradients of their \gls{sde}-induced generative model. \citet{jia2019neural} allow for discontinuities in the latent state to mimic stochastic events. \citet{DengCBML20} and \citet{DengBML21} use dynamic normalizing flows to map the latent state to a continuous-path observation sequence. \citet{kidger2021neural} fit neural \glspl{sde} in a generator-discriminator framework. Like \gls{cru}, these methods accommodate noise in the latent process, but generally rely on variational approximations and numerical solvers. In contrast, \gls{cru} propagates the latent state in closed form and can be trained end-to-end.

%Several authors introduced neural \glspl{sde} as stochastic equivalents to \glspl{node} to accomodate noise in the dynamics of temporal processes %\citep{li2020scalable,tzen2019neural,hodgkinson2021stochastic,liu2019neural}. 
%\citet{jia2019neural} allow for jumps in the latent state to mimic stochastic events. Latent \gls{sde} \citep{li2020scalable} uses the stochastic adjoint sensitive method to compute efficent gradients of their \gls{sde}-induced latent variable model.
%\cite{kidger2021neural} \citet{DengCBML20} and \citet{DengBML21} govern the latent state evolution with neural \glspl{sde} and use dynamic normalizing flows to map to an output sequence. These methods rely on variational approximations and numerical solvers, whereas the state evolution of the \gls{cru} has analytical solutions.

%%%%%%%%%%% model scheme %%%%%%%%%%%%
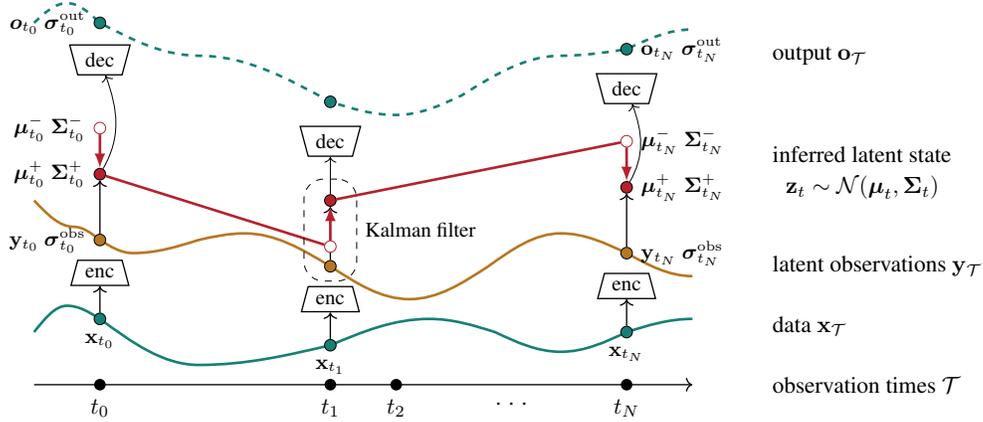
\begin{figure*}
\centering
\resizebox{!}{160pt}{%
\centerline{\begin{tikzpicture}

% time points 
\node[circle,fill=black,inner sep=0pt, minimum size=5pt,label=below:{$t_0$}] (t0) at (1,0) {};
\node[circle,fill=black,inner sep=0pt, minimum size=5pt,label=below:{$t_1$}] (t1) at (4.5,0) {};
\node[circle,fill=black,inner sep=0pt, minimum size=5pt,label=below:{$t_2$}] (t2) at (5.5,0) {};
\node[circle,fill=black,inner sep=0pt, minimum size=5pt,label=below:{$t_N$}] (tn) at (9,0) {};
\node[circle,fill=white,inner sep=0pt, minimum size=8pt,label=below:{. . .}] (tn) at (7.25,0) {};
\node[circle, draw=white,fill=white,inner sep=0pt, minimum size=5pt,label=right:{ observation times $\mathcal{T}$}] (t) at (11,0) {};
%axis
\draw[->, line width=0.2mm] (0,0) -- (10,0);

% observations 	23, 127, 117 
\draw[-, line width=0.4mm, draw=greenish] (0,0.8) sin (0.5,1.2) cos (1,1) sin (2.5,0.3) cos (4.5,0.6) sin (6,1) cos (7, 0.8) sin (8,0.5) cos (9,0.8)  sin (10,1) ;
\node[circle, draw=black, fill=greenish,inner sep=0pt, minimum size=5pt,label=below:{\small $\mathbf{x}_{t_0}$}] (x0) at (1,1) {};
\node[circle, draw=black,fill=greenish,inner sep=0pt, minimum size=5pt,label=below:{\small $\mathbf{x}_{t_1}$}] (x1) at (4.5,0.6) {};
%\node[circle, draw=black,fill=greenish,inner sep=0pt, minimum size=5pt,label=below:{\small $\mathbf{x}_{t_2}$}] (x2) at (5.5,0.95) {};
\node[circle, draw=black,fill=greenish,inner sep=0pt, minimum size=5pt,label=below:{\small $\mathbf{x}_{t_N}$}] (xn) at (9,0.8) {};
\node[circle, draw=white,fill=white,inner sep=0pt, minimum size=5pt,label=right:{data $\mathbf{x}_\mathcal{T}$}] (x) at (11,0.9) {};

% encoders
\node[draw, trapezium, trapezium angle=75, line width=0.2mm ] (enc0) at (1,1.7)  {\small enc};
\node[draw, trapezium, trapezium angle=75, line width=0.2mm ] (enc1) at (4.5,1.3)  {\small enc};
\node[draw, trapezium, trapezium angle=75, line width=0.2mm  ] (encn) at (9,1.5)  {\small enc};

% arrows
\draw[->, line width=0.2mm] (x0) -- (enc0);
\draw[->, line width=0.2mm] (x1) -- (enc1);
\draw[->, line width=0.2mm] (xn) -- (encn);

% latent observations
\draw[-, line width=0.4mm, draw=yellowish] (0,2.8) sin (0.6,2.4) cos (1,2.2) sin (1.5,2) cos (2.5,2.2) sin (3.2, 2.3) cos (4.5,1.8) sin (5.7,1.3) cos (6.9, 1.8) sin (8,2.3) cos (9,2)  sin (10,1.65);
\node[circle, draw=black, fill=yellowish,inner sep=0pt, minimum size=5pt,label=left:{\small $\mathbf{y}_{t_0}$ $\boldsymbol{\sigma}_{t_0}^{\mathrm{obs}}$}] (y0) at (1,2.2) {};
\node[circle, draw=black,fill=yellowish,inner sep=0pt, minimum size=5pt,label=above:{}] (y1) at (4.5,1.8) {};
\node[circle, draw=black,fill=yellowish,inner sep=0pt, minimum size=5pt,label= right:{\small $\mathbf{y}_{t_N}$ $\boldsymbol{\sigma}_{t_N}^{\mathrm{obs}}$}] (yn) at (9,2) {};
\node[circle, draw=white,fill=white,inner sep=0pt, minimum size=5pt,label=right:{latent observations $\mathbf{y}_\mathcal{T}$}] (y) at (11,1.8) {};

% posterior
\node[circle, draw=black, fill=redish,inner sep=0pt, minimum size=5pt,label=left:{\small $\boldsymbol{\mu}_{t_0}^+$ $\mathbf{\Sigma}_{t_0}^{+}$}] (post0) at (1,3.2) {};
\node[circle, draw=black,fill=redish,inner sep=0pt, minimum size=5pt,label=left:{}] (post1) at (4.5,2.8) {};
\node[circle, draw=black,fill=redish,inner sep=0pt, minimum size=5pt,label=right:{\small $\boldsymbol{\mu}_{t_N}^+$ $\mathbf{\Sigma}_{t_N}^{+}$}] (postn) at (9,3) {};
\node[circle, draw=white,fill=white,inner sep=0pt, minimum size=5pt,label=right:{inferred latent state}] (z) at (11,3.5) {};
\node[circle, draw=white,fill=white,inner sep=0pt, minimum size=5pt,label=right:{ $\mathbf{z}_t \sim \mathcal{N}(\boldsymbol{\mu}_t, \boldsymbol{\Sigma}_t)$}] (z) at (11.2,3) {};

% arrows update
\draw[->, line width=0.2mm] (y0) -- (post0);
\draw[->, line width=0.2mm] (y1) -- (post1);
\draw[->, line width=0.2mm] (yn) -- (postn);

% prior
\node[circle, draw=redish, fill=white,inner sep=0pt, minimum size=5pt,label=left:{\small$\boldsymbol{\mu}_{t_0}^-$ $\mathbf{\Sigma}_{t_0}^{-}$}] (prior0) at (1,3.9) {};
\node[circle, draw=redish, fill=white,inner sep=0pt, minimum size=5pt,label=above:{}] (prior1) at (4.5,2.1) {};
\node[circle, draw=redish, fill=white,inner sep=0pt, minimum size=5pt,label=right:{\small $\boldsymbol{\mu}_{t_N}^-$ $\mathbf{\Sigma}_{t_N}^{-}$}] (priorn) at (9,3.7) {};

% update
\draw[-latex, line width=0.4mm,  draw=redish] (prior0) -- (post0);
\draw[-latex, line width=0.4mm,  draw=redish] (prior1) -- (post1);
\draw[-latex, line width=0.4mm,  draw=redish] (priorn) -- (postn);

% prediction
\draw[-, line width=0.4mm, draw=redish] (prior1) -- (post0);
\draw[-, line width=0.4mm, draw=redish] (priorn) -- (post1);

% Kalman Filter
\node[dashed, draw, rounded corners=3mm, minimum width=0.8cm, minimum height = 1.55cm,] (kf) at (4.5, 2.35) {};
\node[anchor=west] at (kf.east)
    {\small Kalman filter};

%decoder
\node[draw, trapezium, trapezium angle=-75, line width=0.2mm, anchor=south ] (dec0) at (1,4.7)  {\small dec };
\node[draw, trapezium, trapezium angle=-75, line width=0.2mm ] (dec1) at (4.5,3.7)  {\small dec};
\node[draw, trapezium, trapezium angle=-75, line width=0.2mm  ] (decn) at (9,4.5)  {\small dec};

% arrow 
\draw (post0) edge[in=-70, out=60, ->] (dec0);
\draw (post1) edge[in=-90, out=90, ->] (dec1);
\draw (postn) edge[in=-70, out=60, ->] (decn);

% filtered observations
\draw[-,dashed, line width=0.4mm, draw=greenish] (0,5.5) sin (0.5,5.8) cos (1,5.5) sin (2.5,5) cos (4.5,4.3) sin (5.5,4.1) cos (6.5, 4.4) sin (8,5) cos (9,5.1)  sin (10,5.4) ;
\node[circle, draw=black, fill=greenish, inner sep=0pt, minimum size=5pt,label=left:{\small $\boldsymbol{o}_{t_0}$ $\boldsymbol{\sigma}_{{t_0}}^{\mathrm{out}}$}] (prior0) at (1,5.5) {};
\node[circle, draw=black, fill=greenish, inner sep=0pt, minimum size=5pt,label=above:{}] (prior1) at (4.5,4.3) {};
\node[circle, draw=black, fill=greenish, inner sep=0pt, minimum size=5pt,label=right:{\small $\mathbf{o}_{t_N}$ $\boldsymbol{\sigma}_{{t_N}}^{\mathrm{out}}$}] (priorn) at (9,5.1) {};
\node[circle, draw=white,fill=white,inner sep=0pt, minimum size=5pt,label=right:{output $\mathbf{o}_{\mathcal{T}}$}] (z) at (11,5) {};
\end{tikzpicture}}}
\vspace{-5pt}
\caption{\gls{cru}: An encoder maps the observation $\mathbf{x}_t$ to a latent observation $\mathbf{y}_t$ and elementwise uncertainties $\boldsymbol{\sigma}^{\mathrm{obs}}_t$. Both are combined with the latent state prior  $\mathcal{N}(\boldsymbol{\mu}_t^- , \mathbf{\Sigma}_t^{-})$ to produce the posterior $\mathcal{N}(\boldsymbol{\mu}_t^+ ,\mathbf{\boldsymbol{\Sigma}_t^{+}})$ (red arrows). A decoder yields the output $\mathbf{o}_t$.}

\label{fig1}
\end{figure*}

\paragraph{Transformers for Irregular Time Series} Besides recurrent and differential equation-based architectures, recent work proposed attention-based methods \citep{vaswani2017attention} to model sequences with arbitrary timestamps. \citet{zhang2019attain} combines time gap decay with attention mechanisms to weight for elapsed time. \citet{horn2020set} use set functions to compress irregular sequences to fixed-length representations.  \Gls{mtand} \citep{ShuklaM21} feeds time embeddings to an attention mechanism. These models are typically quite large, whereas the \gls{cru} achieves high performance despite its small model size.

% CNN method: ICML 2020 \citep{li2020learning}

\section{Method}
\label{sec:method}

The \gls{cru} is a \gls{rnn} for processing sequential data with irregular observation times. It employs a nonlinear mapping (a neural network encoder and decoder) to relate individual observations with a latent state space. In this latent state space, it assumes a continuous latent state whose dynamics evolve according to a linear \gls{sde}. The recursive computations of the \gls{cru} can be derived using the continuous-discrete Kalman filter \citep{jazwinski1970stochastic} and are in closed form. As a result, the \gls{cru} has temporal continuity between hidden states and a gating mechanism that can optimally integrate noisy observations at arbitrary observation times.

We first specify the modeling assumptions for the continuous latent state of the \gls{cru} as well as the role of the encoder and the decoder in \Cref{sec:overview}. In \Cref{sec:kalman}, we derive the 
%predict and update step for the 
recursive internal computations of the \gls{cru} with the resulting recurrent architecture summarized in \Cref{sec:CRU}.
We then develop an efficient \gls{cru} parameterization scheme that affects modeling flexibility and run time (\Cref{sec:params}). Finally, in \Cref{sec:training}, we describe how to train a \gls{cru}.

\subsection{Overview of Proposed Approach}
\label{sec:overview}
The \gls{cru} addresses the challenge of modeling a time series $\mathbf{x}_{\mathcal{T}} = [\mathbf{x}_{t} | t \in \mathcal{T} = \{t_0, t_1, \cdots t_N\}]$ whose observation times $\mathcal{T} = \{t_0, t_1, \cdots t_N\}$ can occur at irregular intervals.
\paragraph{Modeling Assumptions for the Latent State} Unlike the discrete hidden state formulation of standard \glspl{rnn}, the latent state $\mathbf{z} \in \mathbb{R}^M$ of a \gls{cru} has continuous dynamics, which are governed by a linear \gls{sde}
\begin{equation}
\label{eqn:dynamics1}
d\mathbf{z} = \mathbf{A}\mathbf{z}dt  + \mathbf{G}d \boldsymbol{\beta},
\end{equation}
with time-invariant transition matrix $\mathbf{A} \in \mathbb{R}^{M \times M}$ and diffusion coefficient $\mathbf{G} \in \mathbb{R}^{M \times B}$. The integration variable $\boldsymbol{\beta} \in \mathbb{R}^B$ is a Brownian motion process with diffusion matrix $\mathbf{Q} \in \mathbb{R}^{B \times B}$.
The \gls{cru} assumes a Gaussian observation model $\mathbf{H} \in \mathbb{R}^{D \times M}$ that generates noisy latent observations 
\begin{equation}
\label{eqn:dynamics2}
\mathbf{y}_{t} \sim \mathcal{N}(\mathbf{H}\mathbf{z}_{t}, (\boldsymbol{\sigma}_t^{\mathrm{obs}})^2\mathbf{I}),
\end{equation} with observation noise $\boldsymbol{\sigma}_t^{\mathrm{obs}}$.

\paragraph{Sequential Processing}
At each time point $t \in \mathcal{T}$, the latent observation $\mathbf{y}_t$ and its elementwise latent observation noise $\boldsymbol{\sigma}_t^{\mathrm{obs}}$ are produced by a neural network encoder $f_\theta$,
\begin{equation}
    \text{encoder:} \qquad
    \label{eqn:rkn1}
    [\mathbf{y}_t, \boldsymbol{\sigma}_t^{\mathrm{obs}}] = f_{\theta}(\mathbf{x}_t),
\end{equation}
applied to the observation $\mathbf{x}_t$.
At each observation time, we distinguish between a prior and a posterior distribution on $\mathbf{z}_t$.\footnote{We use the notation $\mathbf{y}_{<t} := \{\mathbf{y}_{t^\prime} \,\, \text{for} \,\, t^\prime \in \mathcal{T} \,\, \text{s.t.} \,\, t^\prime <t \}$ for the set of all latent observations before $t$ and $\mathbf{y}_{\leq t} := \{\mathbf{y}_{t^\prime} \,\, \text{for} \,\, t^\prime \in \mathcal{T} \,\, \text{s.t.} \,\, t^\prime \leq t \}$ for this set including $\mathbf{y}_t$.}
\begin{align}
\label{eqn:prior}
    \text{prior:}
    \qquad &p(\mathbf{z}_{t} | \mathbf{y}_{<t}) = \mathcal{N}(\boldsymbol{\mu}^{-}_{t}, \mathbf{\Sigma}^{-}_{t}) \\
    \text{posterior:}  \qquad &p(\mathbf{z}_{t}  | \mathbf{y}_{\leq t}) = \mathcal{N}(\boldsymbol{\mu}^{+}_{t}, \mathbf{\Sigma}^{+}_{t}).
    \label{eqn:posterior}
\end{align}
The parameters of the prior, $\boldsymbol{\mu}^{-}_{t}$ and $ \mathbf{\Sigma}^{-}_{t}$, are computed by propagating the latent state according to \Cref{eqn:dynamics1} (we call this the ``prediction step'') while the parameters of the posterior, $\boldsymbol{\mu}^{+}_{t}$ and $\mathbf{\Sigma}^{+}_{t}$, are computed with a Bayesian update (which we call the ``update step''). The optimal prediction and update step will be derived in closed form in \Cref{sec:kalman}.% using the {\em continuous-discrete Kalman filter} \citep{jazwinski1970stochastic}.

\Cref{fig1} gives an overview of the \gls{cru} from a Kalman filtering perspective: observations (green) are mapped by the encoder into a latent observation space (orange). The mean and variance of the latent state is inferred using the predict and update step of the continuous-discrete Kalman filter (red). 
Finally, a decoder maps the posterior parameters to the desired output space along with elementwise uncertainties. %(dashed green). 
\begin{equation}
\text{decoder:}  \qquad 
\label{eqn:rkn3}
[\mathbf{o}_{t}, \boldsymbol{\sigma}_{t}^{\mathrm{out}}] = g_{\phi}(\mathbf{\boldsymbol{\mu}}_{t}^+, \mathbf{\Sigma}_{t}^{+}).
\end{equation}

%The optimal expression for the parameters in \Cref{eqn:prior,eqn:posterior} will be derived in \Cref{sec:kalman} using a continuous-discrete formulation of the Kalman filter. In \Cref{sec:Then we introduce design choices for \gls{cru} to ensure flexible yet fast state propagation.

%It assumes a continuous latent state $\mathbf{z}\in\mathbb{R}^M$ whose dynamics are governed by a linear \gls{sde}. Figure \ref{fig1} illustrates the network architecture: An encoder and decoder relate observation space with a latent state space. A continuous-discrete Kalman filter alternates between observation updates and continuous state propagation. We first describe the continuous-discrete Kalman filter for the latent state (\Cref{sec:kalman}) and then develop the \gls{cru} in \Cref{sec:CRKN}.

\subsection{Continuous-Discrete Kalman Filter}
\label{sec:kalman}
%The continuous-discrete Kalman filter, see \citet{jazwinski1970stochastic}, assumes a continuous latent state $\mathbf{z} \in \mathbb{R}^M$ that evolves according to the \gls{sde}
The continuous-discrete Kalman filter \citep{jazwinski1970stochastic} is the optimal state estimator for a continuous state space model (\Cref{eqn:dynamics1}) with a discrete-time Gaussian observation process (\Cref{eqn:dynamics2}).  %The latent state $\mathbf{z}$ is assumed to evolve continuously according to the linear \gls{sde} in \Cref{eqn:dynamics1}. At times $t \in \mathcal{T}$ an observation $\mathbf{y}_t \in \mathbb{R}^D$ is sampled from the latent state according to \Cref{eqn:dynamics2}.
This version of the Kalman filter allows modelling observations of a continuous process at potentially arbitrary but discrete observation times. Given the latent observations, the posterior distribution of the latent state (\Cref{eqn:posterior}) is computed recursively, alternating between a predict and an update step. These steps are derived next.

%The posterior computation at observation time $t \in \mathcal{T}$ assumes that we have already computed the posterior at the last observation whose time index is given by

%The prediction step uses the posterior of $\mathbf{z}_{\tau(t)}$ and the \gls{sde} in \Cref{eqn:dynamics1} to produce a prior distribution on $\mathbf{z}_t$. The prior and the latent observation $\mathbf{y}_t$ are combined in the update step to produce the new posterior. This recursion is depicted in \Cref{fig:cell}. %We next derive the prediction and the update step.

\begin{figure}[t]

\begin{tikzpicture}[
    % GLOBAL CFG
    font=\small,
    >=LaTeX,
    % Styles
    cell/.style={% For the main box
        rectangle, 
        rounded corners=5mm, 
        draw,
        very thick,
        },
    operator/.style={%For operators like +  and  x
        circle,
        draw,
        inner sep=-0.5pt,
        minimum height =.2cm,
        },
    function/.style={%For functions
        ellipse,
        draw,
        inner sep=1pt
        },
    ct/.style={% For external inputs and outputs
        },
    gt/.style={% For internal inputs
        rectangle,
        draw,
        minimum width=4mm,
        minimum height=3mm,
        inner sep=1pt
        },
    mylabel/.style={% something new that I have learned
        font=\scriptsize\sffamily
        },
    Arrow/.style={% Arrows with rounded corners
        rounded corners=.35cm,
        thick,
        },
    ]

% CELL
\centering
\node [cell, minimum height =3.5cm, minimum width=6cm] at (0,0){} ;

% OPERATIONS
\node [gt, minimum width=1.2cm, minimum height = 0.5cm] (pred) at (-1,0) {predict};
\node [gt, minimum width=1.2cm, minimum height = 0.5cm] (upd) at (1,0) {update};
\node [gt, minimum width=1.2cm, minimum height = 0.5cm] (enc) at (-2,-1) {encode};
\node [gt, minimum width=1.2cm, minimum height = 0.5cm] (dec) at (2,1) {decode};
\node [ct] (zin) at (-3.5,0) {};
\node [ct] (zout) at (3.7,0) {};
\node [ct] (xin) at (-2,-2.5) {$\mathbf{x}_t$};
\node [ct] (xout) at (2,2.5) {$\mathbf{o}_t$, $\boldsymbol{\sigma}_t^{\mathrm{out}}$};
\node [ct] (tau) at (-0.8,2.1) {$\tau(t) := \max\{t^\prime \in \mathcal{T} \,\, \text{s.t.} \,\, t' < t \}$};

\draw [->, Arrow] (zin) -- (pred) node[midway,above] {$\boldsymbol{\mu}_{\tau(t)}^+$} node[midway,below] {$\boldsymbol{\Sigma}_{\tau(t)}^+$};
  
\draw [->, Arrow] (enc) -| (upd)
node[right]  at (-0.7,-1.3) {$\mathbf{y}_{t}$, $\boldsymbol{\sigma}_t^{\mathrm{obs}}$};
  
\draw [->, Arrow] (pred) -- (upd) node[midway,above] {$\boldsymbol{\mu}_{t}^-$} node[midway,below] {$\boldsymbol{\Sigma}_{t}^-$};
    
\draw [->, Arrow] (upd) -- (zout) node[midway,above] {$\boldsymbol{\mu_{t}}^+$} node[midway,below] {$\boldsymbol{\Sigma_{t}}^+$};
    
\draw [->, Arrow] (upd) -| (dec);

\draw [->, Arrow] (xin) -- (enc);

\draw [->, Arrow] (dec) -- (xout);

\end{tikzpicture}

\caption{ %At each time step $t \in \mathcal{T}$ an encoder and a decoder relate the observation with a latent observation and the internal hidden state with an observed output. 
The internal hidden states of a \gls{cru} cell are the posterior mean and variance $\mu_{t}^+$ and $\Sigma_{t}^+$ of the continuous state variable $\mathbf{z}$. They are computed recursively according to \Cref{alg:cru}.
%based on the posterior parameters at the last observation time $\tau(t)$ using the predict (\Cref{eqn:prior6}) and update step (\Cref{eqn:posterior4}.) The Bayesian treatment of the update step results in a gating mechanism which optimally balances information encoded in previous states with the information gain associated with the incoming observation.
}\label{fig:cell}
\end{figure}
\subsubsection{Prediction Step} Between observation times, the prior density describes the evolution of $\mathbf{z}_t$. It is governed by the \gls{sde} in \Cref{eqn:dynamics1}, which has an analytical solution for linear, time-invariant systems as considered here. To compute the prior at time $t$, we assume that the posterior parameters $\boldsymbol{\mu}_{\tau(t)}^+,\mathbf{\Sigma}_{\tau(t)}^+$ at the last observation time,
\begin{align}
    \tau(t) := \max \{t^\prime \in \mathcal{T} \,\, \text{s.t.}\,\, t^\prime < t\},
\end{align}
have already been computed.
The SDE solution at time $t$ is 
\begin{equation}
\label{eqn:prior3}
\nonumber
\mathbf{z}_{t} = \mathbf{exp}\big(\mathbf{A}(t - \tau(t) )\big)\mathbf{z}_{\tau(t)} 
    + \int_{\tau(t)}^{t} \mathbf{exp}\big(\mathbf{A}(t - s)\big)\mathbf{G}d\boldsymbol{\beta}_{s},
\end{equation}
which results in a prior mean and covariance of
\begin{align}
 \label{eqn:prior4}
    \boldsymbol{\mu}^{-}_{t} &= \mathbf{exp}\big(\mathbf{A}(t - \tau(t) )\big)\boldsymbol{\mu}_{\tau(t)}^+    \\
    \label{eqn:prior5}
        \mathbf{\Sigma}^{-}_{t} &=  \mathbf{exp}\big(\mathbf{A}(t - \tau(t))\big)\mathbf{\Sigma}_{\tau(t)}^+ \mathbf{exp}\big(\mathbf{A}(t - \tau(t) )\big)^T \nonumber \\
        + &\int_{\tau(t)}^{t}  \mathbf{exp}\big(\mathbf{A}(t - s)\big) \mathbf{G}\mathbf{Q} \mathbf{G}^T \mathbf{exp}\big(\mathbf{A}(t - s)\big)^T ds, \nonumber
\end{align}
where $\mathbf{exp}(\cdot)$ denotes the matrix exponential. The integral can be resolved analytically using matrix fraction decomposition and the computation is detailed in \Cref{matfracdecomp}. We summarize the prediction step (\Cref{eqn:prior4}) for the parameters of the prior with 
\begin{equation}\label{eqn:prior6}
[\boldsymbol{\mu}^{-}_{t}, \mathbf{\Sigma}^{-}_{t}] = \mathrm{predict}(\boldsymbol{\mu}_{\tau(t)}^+, \mathbf{\Sigma}_{\tau(t)}^+ , t-\tau(t)).
\end{equation}

\begin{algorithm}[t]
    \caption{The \gls{cru}}\label{alg:cru}
    \begin{algorithmic}
    \STATE \textbf{Input: }Datapoints and their timestamps $\{(\mathbf{x}_t, t)\}_{t \in \mathcal{T}}$
    %\STATE \textbf{Trainable Parameters: }Encoder, decoder, transition net weights $\theta, \phi, \psi$, transition noise  $\mathbf{q}$, transition matrices $\{\mathbf{A}_i\}_{i=1...K}$ 
    
    \STATE \textbf{Initialize:} $\boldsymbol{\mu}^{+}_{t_0} = \mathbf{0}, \mathbf{\Sigma}^{+}_{t_0} = 10 \cdot \mathbf{I}$%, $\tau(t) = t_0$
    
    \FOR{observation times $t > t_0 \in  \mathcal{T}$}
        \STATE $\mathbf{y}_t, \boldsymbol{\sigma}_t^{\mathrm{obs}} = f_{\theta}(\mathbf{x}_t)$
        %\Comment{Encoder}
        
        \STATE $\boldsymbol{\mu}^{-}_{t}, \mathbf{\Sigma}^{-}_{t} = \mathrm{predict}(\boldsymbol{\mu}_{\tau(t)}^+, \mathbf{\Sigma}_{\tau(t)}^+ , t - \tau(t)) $
        %\Comment{Continuous state propagation}
        
        \STATE $\boldsymbol{\mu}^{+}_{t}, \mathbf{\Sigma}^{+}_{t} = \mathrm{update} (\boldsymbol{\mu}_{t}^{-}, \mathbf{\Sigma}_{t}^-, \mathbf{y}_t, \boldsymbol{\sigma}_t^{\mathrm{obs}}) $ %\Comment{Probabilistic gating}

        \STATE $\mathbf{o}_t, \boldsymbol{\sigma}_{t}^{\mathrm{out}} = g_{\phi}(\mathbf{\boldsymbol{\mu}}_{t}^+, \mathbf{\Sigma}_{t}^{+}) $
        %\Comment{Decoder}
        
        %\STATE $\tau(t) = t$
    \ENDFOR
    \STATE \textbf{Return: }$\{\mathbf{o}_t, \boldsymbol{\sigma}_{t}^{\mathrm{out}} \}_{t \in \mathcal{T}}$
    \end{algorithmic}
\end{algorithm}

\subsubsection{Update Step} At the time of a new observation $\mathbf{y}_t$, the prior is updated using Bayes' theorem,
\begin{equation}
p(\mathbf{z}_{t}  | \mathbf{y}_{\leq t}) \propto p(\mathbf{y}_{t} | \mathbf{z}_{t}) p(\mathbf{z}_{t}  | \mathbf{y}_{<t}).
\end{equation}
Due to the Gaussian assumption, the posterior is again Gaussian, and its mean and covariance are given by  
\begin{align}
    \label{eqn:posterior1}
    \mathbf{\boldsymbol{\mu}}_{t}^+ &= \mathbf{\boldsymbol{\mu}}_{t}^{-} + \mathbf{K}_{t}(\mathbf{y}_{t} - \mathbf{H}\mathbf{\boldsymbol{\mu}}_{t}^-)\\
 \label{eqn:posterior2}
    \mathbf{\Sigma}_{t}^{+} &= (\mathbf{I} - \mathbf{K}_{t}\mathbf{H})\mathbf{\Sigma}_{t}^{-}.
\end{align}
The updates can be seen as weighted averages, where the Kalman gain $\mathbf{K}_t$ acts as a gate between prior and observation.
It contrasts observation noise with prior uncertainty and is high when observations have a low noise level,
\begin{equation}
    \label{eqn:posterior3}
    \mathbf{K}_{t} = \mathbf{\Sigma}_{t}^{-} \mathbf{H}^T (\mathbf{H}\mathbf{\Sigma}_{t}^{-} \mathbf{H}^T  + \mathbf{\Sigma}^{\mathrm{obs}}_t)^{-1}.
\end{equation}
We summarize the update step as 
\begin{equation}\label{eqn:posterior4}
[\boldsymbol{\mu}^{+}_{t}, \mathbf{\Sigma}^{+}_{t}] = \mathrm{update} (\boldsymbol{\mu}_{t}^{-}, \mathbf{\Sigma}_{t}^-, \mathbf{y}_t, \boldsymbol{\sigma}_t^{\mathrm{obs}}).
\end{equation}

\subsection{Continuous Recurrent Units}
\label{sec:CRU}
A \gls{cru} is a recurrent neural architecture that uses the predict and update step of a continuous-discrete Kalman filter (\Cref{eqn:prior6,eqn:posterior4}) in an encoder-decoder framework (\Cref{eqn:rkn1,eqn:rkn3}) to sequentially process irregularly sampled time series. An overview of a \gls{cru} cell is given in \Cref{fig:cell}. \Cref{alg:cru} summarizes the recursion, which is used by the \gls{cru} cell to update its internal parameters based on sequential inputs and to produce the output sequence.

Even though the derivation assumes a probabilistic latent state $\mathbf{z}$, the internal computations in the \gls{cru} cell are deterministic, in closed form, and amenable to back-propagation. This means that the \gls{cru} can be used (and trained end-to-end) like other recurrent architectures, such as \glspl{lstm} or \glspl{gru} on various sequence modeling tasks. Its advantage compared to these architectures is that the \gls{cru} handles irregular observation times in a principled manner. %The tasks we study in this paper include sequence interpolation, extrapolation and regression and will be detailed in \Cref{sec:tasks}. 
%\gls{cru} embeds the continuous-discrete Kalman filter into an encoder-decoder framework. The evolution of the network's latent state is modeled with the Kalman filter employing locally linear state transitions. \Cref{alg:cru} summarizes the network's internals.

We next describe parameterization choices for \gls{cru} which lead to more expressive modeling capacity (\Cref{sec:locallin}) and faster computation (\Cref{sec:implementation}) of the state equations. Finally, we present in \Cref{sec:training} how to train a \gls{cru}.

\subsection{Flexible and Efficient Parameterization of the \gls{cru}}
\label{sec:params}

The linearity assumption of the continuous-discrete Kalman filter is advantageous, as it leads to optimal closed-form computations. However, it also limits the expressiveness of the model. The idea of \gls{cru} is that the modelling flexibility of the encoder and decoder mitigates this limitation and that the dimensionality of the state space is large enough for a linearly evolving latent state to lead to expressive relationships between input and output sequences. 

On the other hand, the dimensionality of the latent state cannot be too large in practice as it affects the runtime of the matrix inversion in \Cref{eqn:posterior3} and the matrix exponential in \Cref{eqn:prior4}. To address this trade-off between modeling flexibility and efficient computation, we make certain parameterization choices for the \gls{cru}. In \Cref{sec:locallin}, we describe a locally linear transition model, which maintains the closed form updates of \Cref{sec:kalman} while making the model more flexible. In \Cref{sec:implementation}, we develop \gls{fcru}\glsreset{fcru}, a version of the \gls{cru} with a novel parameterization of the transition matrices via their eigendecompositions. The resulting model has less modeling flexibility than the \gls{cru} but is significantly faster to train and amenable to larger state spaces.

\subsubsection{Locally Linear State Transitions}
\label{sec:locallin}
A locally linear transition model increases the modeling flexibility of \gls{cru} while maintaining the closed form computation of the predict and update steps in \Cref{sec:kalman}. Similar approaches have been used in deep Kalman architectures \citep{karl2016deep,fraccaro2017disentangled}. We employ the parameterization strategy of  \citet{becker2019recurrent}  and design the transition matrix $\mathbf{A}_t$ at time $t$ as a weighted average of $K$ parameterized basis matrices. The weighting coefficients $\alpha^{(k)}_t$ for $k\in\{1...K\}$ are obtained from the current posterior mean $\boldsymbol{\mu}_t^+$ by a neural network $w_\psi$ with softmax output,
\begin{equation}
\label{eqn:cru1}
    \mathbf{A}_t = \sum_{k=1}^{K} \alpha^{(k)}_t \mathbf{A}^{(k)},
    \qquad \text{with} \,\,
    \boldsymbol{\alpha}_t = w_\psi (\boldsymbol{\mu}^{+}_t).
\end{equation}
To reduce the number of parameters, each basis matrix consists of four banded matrices with bandwidth $b$. In addition, we assume a diagonal diffusion matrix $\mathbf{Q}$ whose vector of diagonal entries $\mathbf{q}$ is a time-invariant learnable parameter. The diffusion coefficient $\mathbf{G}$ of the \gls{sde} in \Cref{eqn:dynamics1} is fixed at the identity matrix, i.e. $\mathbf{G}=\mathbf{I}$. This is not restrictive as $\mathbf{G}$ only occurs in combination with the learnable parameter vector $\mathbf{q}$ (\Cref{eqn:prior3}), which is unconstrained. %\Cref{alg:cru} summarizes the network's internals.

\subsubsection{Efficient Implementation} %\Gls{fcru}}
\label{sec:implementation}
The runtime of the \gls{cru} is dominated by two operations: the matrix inversion in the computation of the Kalman gain (\Cref{eqn:posterior3}) and the matrix exponential in the prediction step (\Cref{eqn:prior4}). As in \citet{becker2019recurrent}, there is a trade-off between modeling flexibility and runtime when choosing how to parametrize the model. \citet{becker2019recurrent} use certain factorization assumptions on the state covariance $\boldsymbol{\Sigma}_t$ and the observation model $\mathbf{H}$ that increase speed and stability by simplifying the matrix inversion. \Gls{cru} also benefits from these assumptions, which are detailed in \Cref{approximations}. However, the \gls{cru} has an additional computational bottleneck, namely the matrix exponential in \Cref{eqn:prior4}.

In this section, we develop \gls{fcru}, 
a model variant that benefits from an efficient implementation of the prediction step. 
\gls{fcru} bypasses the computation of the matrix exponential by allowing only commutative and symmetric base matrices $\mathbf{A}^{(k)}$ with related eigenspaces. 
%This approach reduces the costly and numerically unstable matrix exponential of
While this limits the modeling flexibility, it reduces the runtime of the matrix exponential from complexity $O(n^3)$ to matrix multiplication and elementwise operations. %To facilitate computations, our variant exploits assumptions on how far the basis matrices in \Cref{eqn:cru1} can differ from each other. In terms of model flexibility, it can be seen as a compromise between strictly linear state propagation on the one side of the spectrum and locally linear transitions with an unconstrained parameter space on the other side. 

To avoid having to compute an eigenvalue decomposition, we directly parameterize the basis matrices $\mathbf{A}^{(k)}$ in terms of their eigenvalues and eigenvectors, which enable a change of basis. In the projected space, the state transitions are diagonal and the matrix exponential simplifies to the elementwise exponential function. By allowing only commutative, symmetric basis matrices, we can ensure that the matrix exponential in the projected space is invariant to the order in which the $\mathbf{A}^{(k)}$ are summed (\Cref{eqn:cru1}).

\begin{figure}[t]
\centering
\includegraphics[width=0.9\columnwidth]{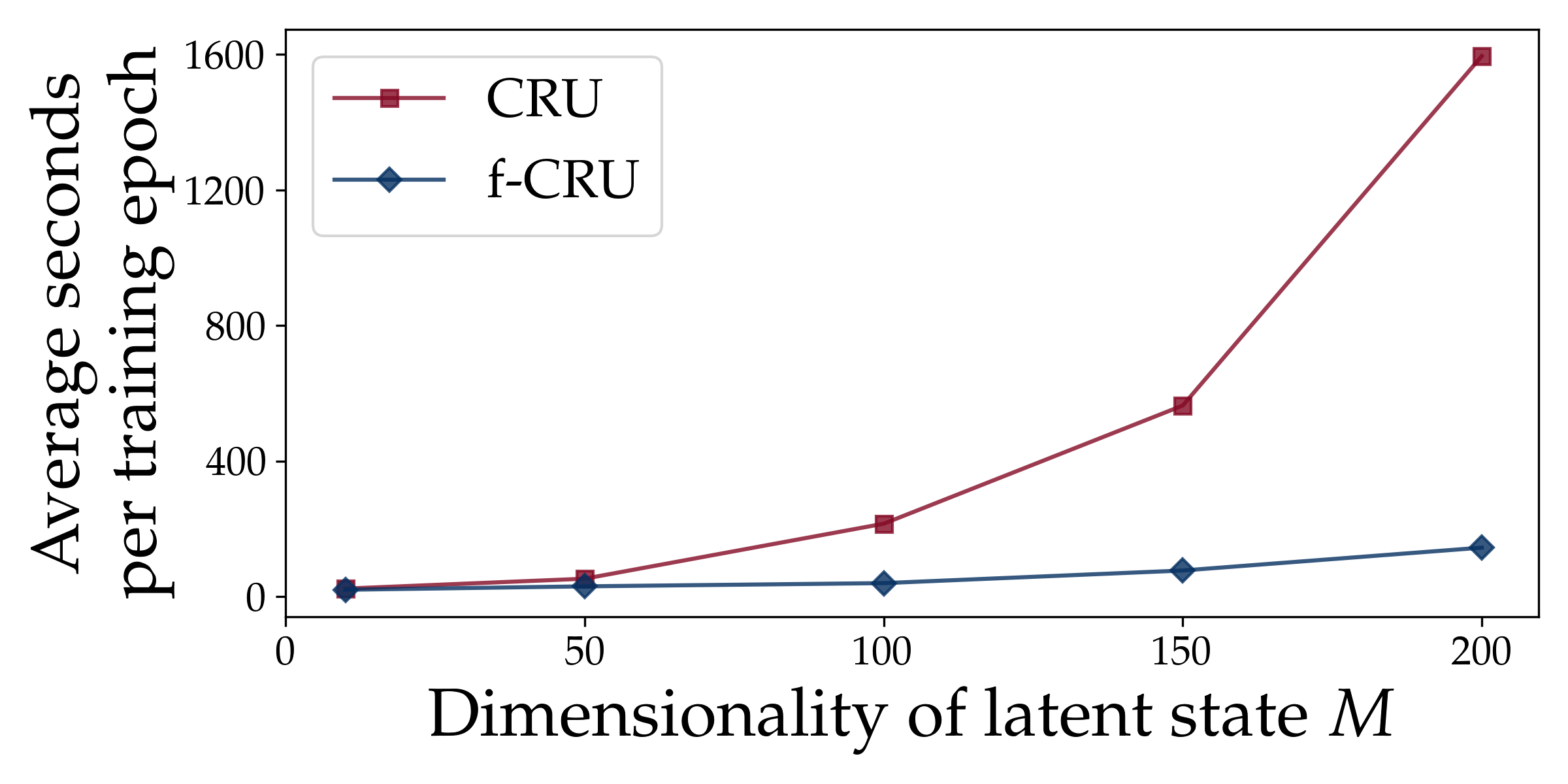}
\caption{The parameterization of the \gls{fcru} makes it faster than the \gls{cru}, especially as the latent dimensionality $M$ increases.}
\label{fig:runtime}
\end{figure}

In detail, we assume diagonalizable basis matrices $\{\mathbf{A}^{(k)}\}_{k=1...K}$ that share the same orthogonal eigenvectors. That is to say, for all $k \in \{1...K\}$ we have $\mathbf{A}^{(k)} = \mathbf{E}\mathbf{D}^{(k)}\mathbf{E}^{T}$ where $\mathbf{D}^{(k)}$ is a diagonal matrix whose $i$-th diagonal entry is the eigenvalue of $\mathbf{A}^{(k)}$ corresponding to the eigenvector in the $i$-th column of $\mathbf{E}$. When using $\mathbf{E}$ to perform a change of basis on the latent state $\mathbf{z}_t$, the \gls{sde} of the transformed state vector $\mathbf{w}$ has diagonal transitions $\mathbf{D}$. The prior mean at time $t$ simplifies to,
\begin{equation}
    \label{eqn:parameterization1}
    \boldsymbol{\mu}^{-}_{t} = \mathbf{E}\exp\Big((t - \tau(t) )\sum_{k=1}^{K}\alpha^{(k)}\mathbf{D}^{(k)}\Big)\mathbf{E}^{T}\boldsymbol{\mu}_{\tau(t)}^+,
    \end{equation}
where $\exp(\cdot)$ denotes the elementwise exponential function. We follow \citet{rome1969direct} to efficiently compute the  covariance of the projected state space $\boldsymbol{\Sigma}_{t}^{\mathbf{w}}$ at time $t$, which is mapped back to the original basis of $\mathbf{z}$ to yield the prior covariance at time $t$
\begin{equation}
    \mathbf{\Sigma}_{t}^{-} = \mathbf{E} \mathbf{\Sigma}_{t}^{\mathbf{w}} \mathbf{E}^{T}
    .% \text{with} \qquad \mathbf{w}_t = \mathbf{E}^{T}\mathbf{z}_t
\end{equation}
We provide the thorough definitions and computations of the \gls{fcru} prior computation in \Cref{covariance}. 
%\gls{fcru} learns the state transitions through the transition's eigenvectors $\mathbf{E}$ and eigenvalues $\{\mathbf{D}^{(k)}\}_{k=1...K}$. 
%While the assumptions on the basis matrices seem restrictive, the free parameters in the encoder and decoder allow for much flexibility to find a suitable state space.
\Cref{fig:runtime} illustrates the computational gain realized by the parameterization scheme of \gls{fcru}. In \Cref{sec:exp}, we further study the speed and accuracy trade-off between  \gls{cru} and \gls{fcru}. 

\subsection{Training}
\label{sec:training}
The trainable parameters of the \gls{cru} are the neural network parameters of the encoder and decoder ($\theta$ and $\phi$), and parameters associated with the locally linear \gls{sde}, namely the diagonal $\mathbf{q}$ of the diffusion matrix, the network parameters $\psi$ for producing the weighting coefficients of the locally linear model, and the transition matrices ($\{\mathbf{A}^{(k)}\}_{i=1...K}$ for the \gls{cru} and $\mathbf{E}$, $\{\mathbf{D}^{(k)}\}_{i=1...K}$ for the \gls{fcru}).   

To train these parameters, we assume a dataset of sequences where each input sequence $\mathbf{x}_\mathcal{T}$ is associated
with a target output sequence $\mathbf{s}_{\mathcal{T}}$. %\footnote{depending on the task, the target output sequence can be a version of the input sequence, e.g. a shifted version for one-step-ahead prediction or an upsampled version for an imputation task.} 
%Depending on the type of target output, the loss is either a Gaussian or a Bernoulli likelihood. 
For real valued outputs, the objective function is the Gaussian negative log-likelihood of the ground truth $\mathbf{s}_{\mathcal{T}}$ and is given for a single sample by 
\begin{equation}
\label{eqn:rkn4}
\mathcal{L}(\mathbf{s}_{\mathcal{T}}) = - \frac{1}{N} \sum_{t \in \mathcal{T}} \mathrm{log}\mathcal{N}(\mathbf{s}_{t}  | {\mathbf{o}_t}, (\boldsymbol{\sigma}_t^{\mathrm{out}})^2 ), 
\end{equation}
where $\mathbf{o}_\mathcal{T}$ and the elementwise uncertainty estimate $\boldsymbol{\sigma}^\mathrm{out}_\mathcal{T}$ is the output computed by the \gls{cru}.
For binary outputs, the model is trained on the Bernoulli negative log-likelihood
\begin{equation}
\label{eqn:rkn5}
\mathcal{L}(\mathbf{s}_{\mathcal{T}}) = - \frac{1}{N} \sum_{t \in \mathcal{T}} \sum_{i=1}^{D_0} s_t^{(i)} \mathrm{log}(o_{t}^{(i)}) +   (1-s_t^{(i)}) \mathrm{log}(1-o_{t}^{(i)}).
\end{equation}
We also use this loss for imputing gray scale images, where the outputs  $\mathbf{s}_t$ are $D_0$ pixel values in the range $[0,1]$. To maintain the orthogonality constraint on $\mathbf{E}$ during training of \gls{fcru}, we use the tools by \citet{lezcano2019trivializations}. 

\iffalse
Because prediction and update step of the continuous-discrete Kalman filter have closed-form solutions, the model operations are deterministic. Thus, like standard \glspl{rnn} the model parameters ($\theta, \phi, \psi, \mathbf{q}$) and  $\{\mathbf{A}^{(k)}\}_{i=1...K}$ or alternatively $\mathbf{E}, \{\mathbf{D}^{(k)}\}_{i=1...K}$ for \gls{fcru} can be trained end-to-end with backpropagation.
\fi

\section{Empirical Study}
\label{sec:exp}
We study the \gls{cru} on three different tasks (interpolation, regression, and extrapolation) on challenging datasets from meteorology and health care. In sequence interpolation (\Cref{sec:interpolation}), the task is to learn the underlying dynamics of the data based on a subset of the observations. In \Cref{sec:regression}, we study a regression task consisting of predicting the angle of a pendulum from noisy images of the system. Finally, in \Cref{sec:extrapolation}, models learn the dynamics of a sequence from the first half of the observations to predict the remaining sequence. This extrapolation is challenging because the models need to capture long-term temporal interactions.

We study \gls{cru} in comparison to other sequence models in terms of both accuracy and runtime. We find that (i) \gls{cru} trains consistently faster than \gls{node}-based models, (ii) our methods outperforms  most  baselines on interpolation and regression, (iii) the uncertainty driven gating mechanisms handles noisy and partially observed inputs systematically by attributing less weight to them in the latent state update. 

\subsection{Datasets}
\paragraph{Pendulum Images} We used the pendulum simulation of \citet{becker2019recurrent} to generate 4 000 synthetic image sequences. The 24x24 pixel images show a pendulum at 50 irregular time steps $\mathcal{T}$. For the interpolation task, half of the frames of each sequence are removed at random, resulting in an input sequence $\mathbf{x}_\mathcal{S}$ with a reduced set of indices $\mathcal{S} \subset \mathcal{T}$. The target output is the full image sequence $\mathbf{s}_\mathcal{T} = \mathbf{x}_\mathcal{T}$.

We also use the pendulum image sequences %$\mathbf{x}_\mathcal{T}$ 
to study \gls{cru} on a regression task. Each frame $\mathbf{x}_t$ is associated with a target label $\mathbf{s}_t = (\textrm{sin}(s_t), \textrm{cos}(s_t))$, which is the angle of the pendulum.  As in \citet{becker2019recurrent}, we corrupt the images with a correlated noise process. We used 2 000 sequences for training and 1 000 for validation and testing each. 
%For both tasks, the sequences are split into a training ($80\%$), a validation ($10\%$) and a test set ($10\%$).

%In the first setting, we remove half of the images per sequence and task the models to impute the missing frames. In the second setting, we keep the entire sequence and add a correlated noise process to the images as in \citet{becker2019recurrent}. The noise increases over consecutive time steps resulting in images potentially consisting of pure noise. Here, we evaluate the models on detecting the pendulum angle  at each time step. 

\paragraph{Climate Data (USHCN)}
The \gls{ushcn} dataset \citep{menne2015long} contains daily measurements from 1 218 weather stations across the US for five variables: precipitation, snowfall, snow depth, minimum and maximum temperature. We used the cleaning procedure by \citet{brouwer2019gru} to select a subset of 1168 meteorological stations over a range of four years (1990 - 1993). Though collected in regular time intervals, climatic data is often sparse due to, e.g., sensor failures. To further increase the sporadicity of the data across time and dimensions, we first subsample 50\% of the time points and then randomly remove 20\% of the measurements. We test models on a 20\% hold-out set and trained on 80\% of which we used 25\% for validation.

\paragraph{Electronic Health Records (Physionet)} Finally, we also benchmark the models on the data set of the Physionet Computing in Cardiology Challenge 2012 \citep{silva2012predicting}. The data reports 41 measurements of the first 48 hours of 8000 ICU patients. We follow the preprocessing of \citet{rubanova2019latent} and round observation times to 6 minutes steps resulting in 72 different time points per patient on average. At a single time point, an observation contains on average only 16\% of the features with measurements entirely missing for some patients. We split the data into 20\% test and 80\% train set of which we used 25\% for validation.

\subsection{Baselines}
We study the \gls{cru} and \gls{fcru} in comparison to various baselines including \gls{rnn} architectures known to be powerful sequence models (but  for regular observation times) as well as \gls{ode} and attention-based models, which have been developed specifically for observations from continuous processes. 
%To make comparison fair, we used the same latent state size across architectures. When processing images, we extend baselines with the convolutional encoder and decoder used by \gls{cru}. Please refer to \Cref{sec:implementation_details} for implementation details.

\paragraph{Recurrent Neural Networks} We compare our method against two \gls{rnn} architectures with discrete hidden state assumption: \glspl{gru} \citep{chung2014empirical} and \glspl{rkn} \citep{becker2019recurrent}. To aid these models with irregular time intervals, we run a version where we feed the time gap $\Delta_t = t - \tau(t)$ as an additional input to the model (denoted as \gls{rkn}-$\Delta_t$ and \gls{gru}-$\Delta_t$).
%For these models, we discretize the timeline in a preprocessing step. Precisely, we split the timeline into 20 equally spaced bins and keep the time point that is closest to the bin center. 
Another baseline is \gls{gru}-D \citep{che2018recurrent}, which uses trainable hidden state decay between observations to handle irregular inputs on a continuous time scale.

\paragraph{ODE-based Models} We also test \gls{cru} against three \gls{ode}-based models that can naturally deal with irregularly sampled time series: (1) ODE-RNN \citep{rubanova2019latent} alternates between continuous hidden state dynamics defined by an \gls{ode} and classical \gls{rnn} updates at observation times.
(2) Latent \gls{ode} (\citet{chen2018neural}, \citet{rubanova2019latent}) is a generative model that uses ODE-RNN as recognition network to infer the initial value of its latent state and models the state evolution with an \gls{ode}. (3) GRU-ODE-B \citep{brouwer2019gru} combines a continuous-time version of \glspl{gru} with a discrete update network.

\paragraph{Attention-based Model} We also compare \Gls{cru} against a transformer network: \gls{mtand} \cite{ShuklaM21} is a generative model that employs multi-time attention modules in the encoder and decoder. 
% \citep{RezendeMW14,KingmaW13} framework. 

\paragraph{Implementation Details}
For a fair comparison, we use the same latent state dimension for all approaches ($M=30$ for pendulum, $M=20$ for Physionet and $M=10$ for USHCN), except for \gls{gru}, where we increase the latent state size such that the number of parameters is comparable to \gls{cru}. For the Physionet and \gls{ushcn} experiments, the encoder and decoder architecture of the \gls{cru} mimic that of the latent ODE in \citet{rubanova2019latent}'s Physionet set-up. (Details can be found in \Cref{implementation_details}.)

For processing pendulum images, the \gls{cru} encoder and decoder are a convolutional architecture (see \Cref{implementation_cru}). To ensure a fair comparison, we follow the  set-up of \citet{becker2019recurrent} and use the same encoder and decoder architecture as for \gls{rkn}, \gls{cru} and \gls{fcru} to give the same feature extraction capabilities to the other baseline models: the baseline models are applied to the latent observations that the encoder produces from the inputs and the decoder maps the baseline outputs to the target output. In this augmented framework, the encoder, the baseline model, and the decoder are trained jointly.
%To scale the methods based on \glspl{node} to the pendulum experiments, we use the same encoder and decoder architecture as for \gls{cru} to extract image features for latent \gls{ode} and \gls{ode}-\gls{rnn}.
More information and other implementation details can be found in \Cref{implementation_details}.

\begin{table*}[ht!]
\caption{Test MSE (mean $\pm$ std) and runtime (average seconds/epoch) for interpolation and extrapolation on \gls{ushcn} and Physionet. }
\vspace{-10pt}
\label{interpolation}
\begin{center}
\begin{tabular}{l@{\hskip 25pt}cc@{\hskip 30pt}cc@{\hskip 20pt}cc}
& \multicolumn{2}{c}{Interpolation MSE ($\times10^{-2}$)} & \multicolumn{2}{c}{Extrapolation MSE ($\times10^{-2}$)} &
\multicolumn{2}{c}{Runtime (sec./epoch)} \\
Model & \gls{ushcn} & Physionet & \gls{ushcn} & Physionet & \gls{ushcn} & Physionet \\
\toprule
mTAND & 1.766 $\pm$ 0.009 & 0.208 $\pm$ 0.025 & 2.360	$\pm$ 0.038 & \textbf{0.340 $\pm$ 0.020} & 7 & 10 \\
RKN & 0.021	$\pm$ 0.015 & 0.188 $\pm$ 0.088& 1.478 $\pm$ 0.283
 & 0.704 $\pm$ 0.038
& 94 & 39 \\
RKN-$\Delta_t$ & \textbf{0.009 $\pm$ 0.002}
& 0.186 $\pm$ 0.030
& 1.491 $\pm$ 0.272
 & 0.703 $\pm$ 0.050
& 94 & 39\\
GRU & 0.184 $\pm$ 0.183
& 0.364 $\pm$ 0.088
& 2.071 $\pm$ 0.015
& 0.880 $\pm$ 0.140
& 3 & 5\\
GRU-$\Delta_t$ & 0.090 $\pm$ 0.059
& 0.271 $\pm$ 0.057
& 2.081 $\pm$ 0.054
& 0.870 $\pm$ 0.077
& 3 & 5 \\
GRU-D  & 0.944 $\pm$ 0.011   & $0.338^*$ $\pm$ 0.027 &   $1.718 \pm 0.015$ & 0.873 $\pm$ 0.071 &   292 &  5736\\
Latent ODE  & 1.798 $\pm$ 0.009  & $0.212^*$ $\pm$ 0.027  &   $2.034 \pm 0.005$ & 0.725 $\pm$ 0.072 & 110  & 791 \\
ODE-RNN    & 0.831 $\pm$ 0.008   & $0.236^*$ $\pm$ 0.009  &  $1.955 \pm 0.466$ & 0.467 $\pm$ 0.006 & 81 &  299 \\
GRU-ODE-B  & 0.841 $\pm$ 0.142     & 0.521 $\pm$ 0.038  &  $5.437 \pm 1.020$ & 0.798 $\pm$ 0.071  & 389 & 90\\
\gls{fcru} (ours)   & 0.013 $\pm$ 0.004   & 0.194 $\pm$ 0.007  &  $1.569 \pm 0.321$ & 0.714 $\pm$ 0.036 &  61 & 62 \\
\Gls{cru} (ours) & 0.016 $\pm$ 0.006 & \textbf{0.182} $\pm$ \textbf{0.091}  &  \textbf{1.273} $\pm$  \textbf{0.066} & 0.629 $\pm$ 0.093 &  122   & 114 \\
\toprule
\end{tabular}
\end{center}
\vspace{-5pt}
$^*$ \small Results from \citet{rubanova2019latent}.
\end{table*}

\subsection{Results on Sequence Interpolation}
\label{sec:interpolation}
We first examine the efficacy of \gls{cru} in sequence interpolation on pendulum images, \gls{ushcn} and Physionet. The task consists of inferring the entire sequence $\mathbf{s}_\mathcal{T}=\mathbf{x}_{\mathcal{T}}$ based on a subset of observations $\mathbf{x}_{\mathcal{S}}$ where $\mathcal{S} \subseteq \mathcal{T}$. In the pendulum interpolation task, $\mathcal{S}$ contains half of the total time points sampled at random. For \gls{ushcn} and Physionet, we follow the Physionet set-up in \citet{rubanova2019latent}, where reconstruction is based on the entire time series, i.e. $\mathcal{S} = \mathcal{T}$. %Models are trained to reconstruct the entire sequence $\mathbf{s}_\mathcal{T}=\mathbf{x}_{\mathcal{T}}$. 

\Cref{interpolation} reports \gls{mse} on full test sequences and average runtime per epoch for \gls{ushcn} and Physionet. \Cref{pendulum} summarizes results on the pendulum image imputation task. All reported results are averages over 5 runs.  The runtime measures a training pass through the entire dataset while keeping settings such as batch size and number of parallel threads comparable across architectures. Both \gls{cru} and \gls{fcru} outperform baseline models on the interpolation task on most datasets. The efficient implementation variant, \gls{fcru}, is consistently faster than \gls{node}-based architectures. \gls{fcru} produces results comparable to the \gls{cru} baseline, while reducing training time by up to 50\% even in low-dimensional state spaces. Though discrete \glspl{rnn} are still faster, \gls{fcru} takes only a fraction of time to train as neural \gls{ode}-based methods. 

Three noteworthy factors influence training and inference time: (1) Unlike \gls{ode}-based methods, \gls{cru} has closed-from computation, whereas \gls{ode}-based methods still need to call an ODE solver during inference.
(2) Our method can make efficient use of batching and unlike \gls{node}-based architectures, \gls{cru} does not require solving the state propagation in sync for the union of all timestamps in a minibatch. Thus, the complexity of \gls{cru} does not scale with the heterogeneity of timestamps in a minibatch. On data where timestamps vary widely across sequences (e.g. Physionet), the number of required update steps can be up to $B$-times more for recurrent \gls{node}-architectures than for \gls{cru}, where $B$ denotes batch size. A batch size of 1 foregoes this advantage of CRU. (3) The operations of the encoder and decoder can counteract the speedup gained from closed-form latent state propagation. We found the speedup to be less significant on the pendulum data, where the encoder and decoder have a stronger influence on runtime (\Cref{pendulum}).

\subsection{Results on Pendulum Angle Prediction from Images}
\label{sec:regression}
Next, we consider a regression task on irregularly sampled pendulum image sequences. Here, each  observed image $\mathbf{x}_t$ is mapped to a continuous target variable $\mathbf{s}_t$ representing the pendulum angle at each time step. To assess the noise robustness of the methods, the image sequences are corrupted by a correlated noise process as in \citet{becker2019recurrent}.

\begin{figure*}[t]
    \centering
    \includegraphics[width=\textwidth]{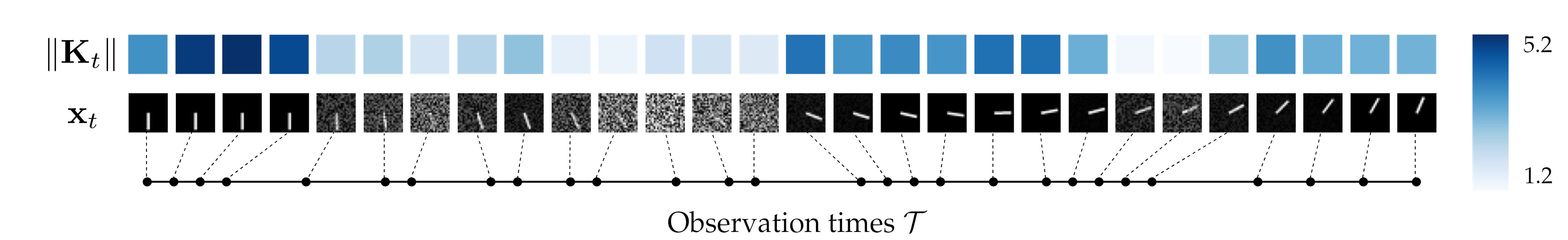}
    %\vskip -0.1in
    \caption{Pendulum angle prediction: The figures show noise corrupted observations and the corresponding Kalman gain norm for a pendulum trajectory sampled at irregular time intervals. The Kalman gain reflects the underlying noise process of the data.}
    \label{fig:noise}
    \vskip -0.1in
\end{figure*}

Figure \ref{fig:noise} shows how the gating mechanism of \glspl{cru} works under varying degrees of observation noise. The norm of the Kalman gain mirrors the noise process of the sample sequence. At times of high observation noise, the norm of the Kalman gain is small and consequently the state update is dominated by the history encoded in the latent state prior. In contrast, when the pendulum is clearly observed, the Kalman gain attributes high weight to the new observation. 
%Second, we observe that the Kalman gain norm at time $t$ is relative to the noise level of the previous image. When the previous image $\mathbf{x}_{\tau(t)}$ is clearly observed, a light increase of the noise level in $\mathbf{x}_t$ can decrease the importance of $\mathbf{x}_t$ for the update sharply.

This principled handling of noise is one of the factors that can help explain the success of \gls{cru} in pendulum angle prediction.
Results in terms of MSE are shown in \Cref{pendulum}. \Gls{cru} outperforms existing baselines. In \Cref{sec:loglikelihood}, we also report log-likelihood results for this task. \gls{cru} and \gls{fcru} yield the best performance.

\begin{table}
%\footnotesize
\caption[Pendulum interpolation results]{Test \gls{mse} $\times10^{-3}$ (mean $\pm$ std) and runtime (average sec/epoch) on pendulum interpolation and regression.}
\begin{center}
\vskip -0.1in
\begin{tabular}{l@{\hskip 5pt}cccc}
Model & Interpolation & {\footnotesize R.time} & Regression   \\
\toprule
%\gls{gru} (binned) & 5.144 $\pm$ 0.057  & 12 \\
mTAND & 15.400 $\pm$ 0.071 & 3 & 65.640 $\pm$ 4.050 \\
%\gls{rkn} (binned)  & 5.026 $\pm$  0.092  & 20 \\
\gls{rkn} & 5.200 $\pm$ 0.051 & 20 & 8.433	 $\pm$ 0.610 \\
\gls{rkn}-$\Delta_t$  & 1.903 $\pm$  0.137  & 20 & 5.092 $\pm$ 0.395\\
\gls{gru}  & 5.086 $\pm$ 0.028 
 & 12 & 9.435	$\pm$ 0.998 \\
\gls{gru}-$\Delta_t$  & 2.073 $\pm$ 0.185 & 12 & 5.439 $\pm$ 0.988  \\
Latent ODE & 15.060 $\pm$ 0.138 & 52 & 15.700 $\pm$ 2.848\\
ODE-RNN & 2.830 $\pm$ 0.137 & 37 & 7.256 $\pm$ 0.406 \\
GRU-ODE-B & 9.144 $\pm$ 0.020 & 60 & 9.783 $\pm$ 3.403\\
\gls{fcru} & 1.386 $\pm$ 0.162 & 29 & 6.155 $\pm$ 0.881 \\
\gls{cru} &  \textbf{0.996} $\pm$  \textbf{0.052}   & 36 &  \textbf{4.626} $\pm$ \textbf{1.072}\\
\toprule \\
\label{pendulum}
\end{tabular}
\end{center}
\vskip -.6in
\end{table}

\subsection{Results on Sequence Extrapolation}
\label{sec:extrapolation}

Finally, we study the performance of \gls{cru} on extrapolating sequences far beyond the observable time frame. We split the timeline into two halves $\mathcal{T}_1 = \{t_0,...t_k\}$ and $\mathcal{T}_2 = \{t_{k+1},...t_N\}$. Models are tasked to predict all time points of the sequence $\mathcal{T} = \mathcal{T}_1 \cup \mathcal{T}_2$ based on time points in $\mathcal{T}_1$ only. In the Physionet experiment, the input consists of the first 24 hours and the target output of the first 48 hours of patient measurements. For \gls{ushcn} we split the timeline into two parts of equal length $t_k = N/2$. During training, the entire observation sequence is used as target
%, i.e. $\mathbf{s}_\mathcal{T}$ = $\mathbf{x}_{\mathcal{T}}$
to guide the training process, except for GRU-ODE-B, where we used the training strategy proposed by the authors. During evaluation, performance is assessed only on the extrapolated part of the test sequences.

\Cref{interpolation} reports test errors on the extrapolated test sequences, $\mathbf{s}_{\mathcal{T}_2}$. On the Physionet data, \gls{mtand} achieves the lowest errors. On climate data, \gls{cru}
%and \gls{fcru} produce better results than the baselines with the flexible transition model of the \gls{cru} 
reaches the highest performance.

%\subsection{Qualitative analysis of the probabilistic state update}
%We are interested in how the probabilistic state space depicts uncertainty when updating the latent state. Two sources of uncertainty are considered: uncertainty arising from noise and from missing features. 

\begin{figure}[t]
%\vskip -0.2in
\centering
\includegraphics[width=0.7\columnwidth]{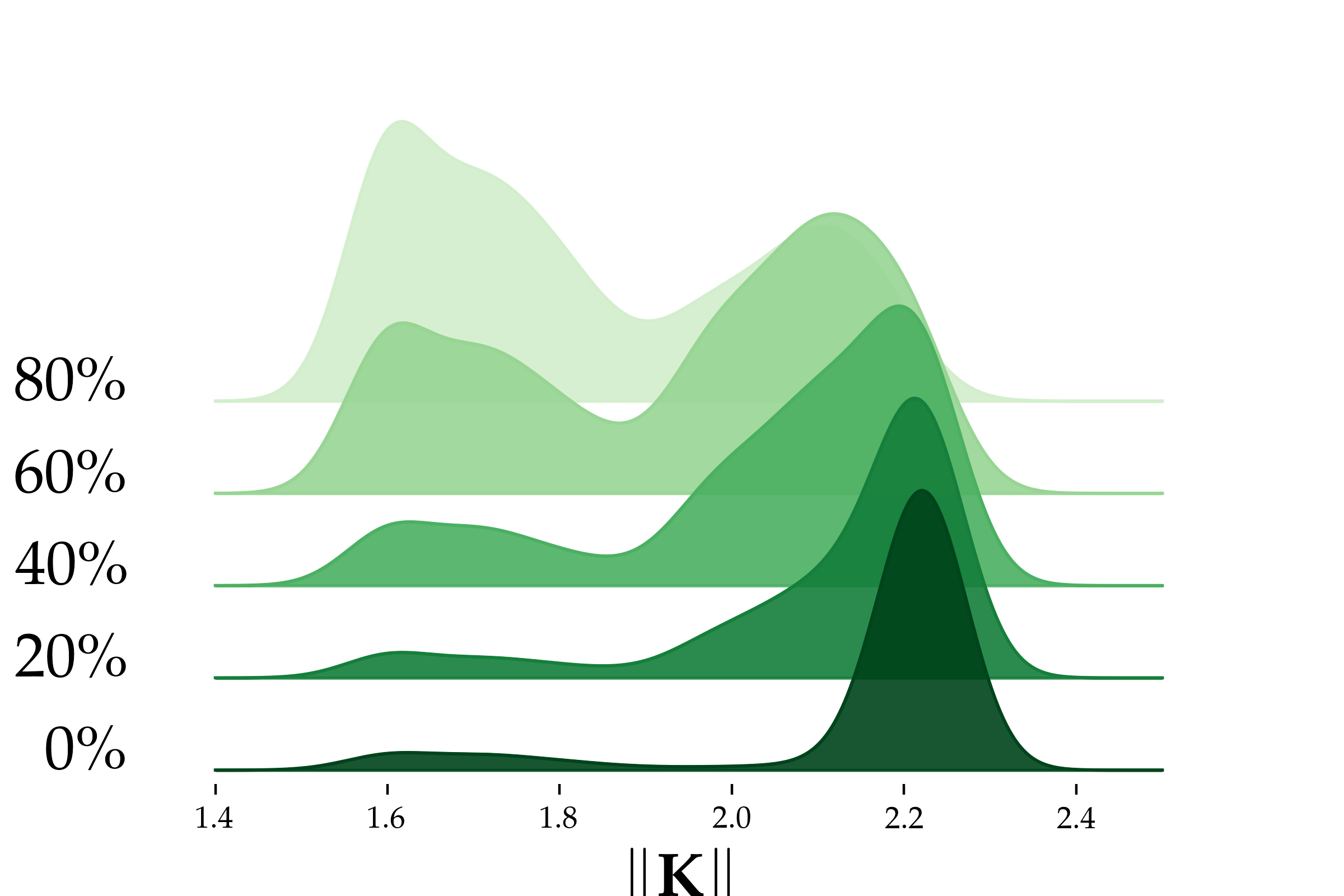}
\caption{\gls{ushcn} extrapolation: Distribution of the Kalman gain norm for different percentages of sparseness. For highly sparse observation vectors (e.g. 80\% sparse), the distribution is shifted towards low norm values.}
\label{fig:sparseness}
\vskip -0.2in
\end{figure}

\paragraph{Partial Observability} To study the \gls{cru} gating mechanism in the presence of partially observed inputs, we also run the extrapolation task on the five-dimensional \gls{ushcn} data while, this time, explicitly controlling the degree of partial observability. For each time point, we sample one to four features each with probability 0.1. The remaining 60\% of observations are fully observed to ensure stable learning. \Cref{fig:sparseness} shows how the Kalman gain reacts to the sparseness of an observation. The distribution of fully observed feature vectors (0\% sparseness) is centered around comparably high values of the Kalman gain norm, whereas sparse observations are associated with lower Kalman gain norm values. Thus, sparse observations are given less weight in the latent state update than fully observed feature vectors.

\section{Conclusion}
We have developed \gls{cru}, a \gls{rnn} that models temporal data with non-uniform time intervals in a principled manner. It incorporates a continuous-discrete Kalman filter into an encoder-decoder structure thereby introducing  temporal continuity into the hidden state and a notion of uncertainty into the network's gating mechanism. Our empirical study finds that the gating mechanism of the \gls{cru} weights noisy and partially observed input data accurately. Our method outperforms established recurrent sequence models such as \gls{gru} on irregularly sampled data and achieves better interpolation accuracy than \gls{node}-based models on challenging datasets from various domains.

\section*{Acknowledgements}
We thank Philipp Becker for providing the RKN code, Klaus Obermayer for helpful discussion, and Stefan Kurz for valuable feedback on the draft.
The Bosch Group is carbon neutral. Administration, manufacturing and research activities do no longer leave a carbon footprint. This also includes GPU clusters on which the experiments have been performed.


%\nocitet{*}
\bibliography{main}
\bibliographystyle{icml2022}

\newpage
\onecolumn
\appendix

\section{Prior Computation}
\subsection{\gls{fcru}}
For an efficient implementation, we assume locally linear transitions with symmetric basis matrices $\{\mathbf{A}^{(k)}\}_{k=1...K}$ that share the same eigenvectors,

\begin{equation}
\label{eqn:transition}
    \mathbf{A}_t = \sum_{k=1}^{K} \alpha^{(k)}_t \mathbf{A}^{(k)},
    \qquad \text{with} \qquad
    \mathbf{A}^{(k)} = \mathbf{E}\mathbf{D}^{(k)}\mathbf{E}^T.
\end{equation}

\subsubsection{Prior Mean}
We can simplify the matrix exponential in \Cref{eqn:prior4} to the elementwise exponential function:

\begin{equation}
    \begin{aligned}
        \boldsymbol{\mu}^{-}_{t} & = \mathbf{exp}\Big(\mathbf{A}_t(t - \tau(t) )\Big)\boldsymbol{\mu}_{\tau(t)}^+ \\
        & = \mathbf{exp}\Big(\sum_{k=1}^{K}\alpha^{(k)}\mathbf{A}^{(k)}(t - \tau(t) )\Big)\boldsymbol{\mu}_{\tau(t)}^+ \\
        & = \mathbf{exp}\Big(\sum_{k=1}^{K}\alpha^{(k)}\mathbf{E}\mathbf{D}^{(k)}\mathbf{E}^{T}(t - \tau(t) )\Big)\boldsymbol{\mu}_{\tau(t)}^+ \\
        & = \mathbf{E}\exp\Big(\sum_{i=1}^{K}\alpha^{(k)}\mathbf{D}^{(k)}(t - \tau(t) )\Big)\mathbf{E}^{T}\boldsymbol{\mu}_{\tau(t)}^+
    \end{aligned}
\end{equation}

In the last step, we exploit a property of the matrix exponential. The exponential of diagonalizable matrices can be obtained by exponentiating each entry on the main diagonal of the matrix of eigenvalues.

\subsubsection{Prior Covariance}
\label{covariance}
For diagonalizable transition matrices of a linear time-invariant system \citet{rome1969direct} proposes an analytical solution for the computation of the prior covariance based on the eigendecomposition of the transition matrix. Precisely, we can define a new state vector $\mathbf{w}$ with covariance $\mathbf{\Sigma}_{\tau(t)}^{\mathbf{w}}$

\begin{equation}
    \mathbf{w}_{\tau(t)} = \mathbf{E}^{T}\mathbf{z}_{\tau(t)} \qquad
    \mathbf{\Sigma}_{\tau(t)}^{\mathbf{w}} =  \mathbf{E}^{T} \mathbf{\Sigma}^+_{\tau(t)} \mathbf{E} 
\end{equation}

We further define $\mathbf{\tilde D}$ as the matrix whose $ij$-th element is the sum of the $i$-th and $j$-th diagonal entry of $\sum_{k=1}^K \mathbf{D}^{(k)}$, 
\begin{equation}
    \mathbf{\tilde D}_{ij} = \sum_{k=1}^K \mathbf{D}_{ii}^{(k)} + \mathbf{D}_{jj}^{(k)}
\end{equation}

Let $\mathbf{S}$ be the transformed noise component,
\begin{equation}
    \mathbf{S} = \mathbf{E}^{T} \mathbf{G}\mathbf{Q}\mathbf{G}^{T} \mathbf{E},
\end{equation} 
then we can compute the covariance of $\mathbf{w}$ at time $t$ with 

\begin{equation}
    \mathbf{\Sigma}_{t}^{\mathbf{w}} = (\mathbf{S} \odot \exp(\mathbf{\tilde D}(t - \tau(t))) - \mathbf{S}) \oslash \mathbf{\tilde D} + \mathbf{\Sigma}_{\tau(t)}^{\mathbf{w}} \odot \exp(\mathbf{\tilde D}(t - \tau(t)))
\end{equation}

Mapping back to the space of $\mathbf{z}$, we obtain for the prior covariance of $\mathbf{z}$ at time $t$

\begin{equation}
    \mathbf{\Sigma}_{t}^{-} = \mathbf{E} \mathbf{\Sigma}_{t}^{\mathbf{w}} \mathbf{E}^{T}
\end{equation}

\subsection{\gls{cru}}
\subsubsection{Prior Covariance}
\label{matfracdecomp}
The integral in the prior covariance (\Cref{eqn:prior4}) can be resolved analytically using matrix fraction decomposition \citep{axelsson2014discrete}. The idea of the method consists in computing the integral by solving a matrix valued ODE. We compute the matrix exponential of the matrix $\textbf{B}$
\begin{equation}
\label{eqn:matfracdecomp1}
\mathbf{B} = \begin{pmatrix}
  \mathbf{A} & \mathbf{GQG}^T\\ 
  \mathbf{0} & -\mathbf{A}^T
  \end{pmatrix}
\qquad
\mathbf{exp}(\mathbf{B}(t-\tau(t))) = \begin{pmatrix}
   \mathbf{M}_{1} & \mathbf{M}_{2} \\ 
   \mathbf{0} & \mathbf{M}_{3} 
   \end{pmatrix}
\end{equation}

where $\mathbf{M}_1$, $\mathbf{M}_2$, $\mathbf{M}_3$ and $\mathbf{0}$ are of dimension $M\times M$. The prior covariance matrix
at time $t$ is then given by

\begin{equation}
\label{eqn:matfracdecomp2}
    \begin{aligned}
        \mathbf{\Sigma}^{-}_{t} = & \mathbf{exp}\big(\mathbf{A}(t - \tau(t))\big)\mathbf{\Sigma}_{\tau(t)}^+ \mathbf{exp}\big(\mathbf{A}(t - \tau(t) )\big)^T \\ & +\mathbf{M}_{2}\mathbf{exp}\big(\mathbf{A}(t - \tau(t) )\big)^T
    \end{aligned}
\end{equation}

\section{Approximations}
\label{approximations}
In their work on \glspl{rkn}, \citet{becker2019recurrent} exploit assumptions on the structure of components of the Kalman filter to simplify computation. In particular, they reduce the matrix inversion in the Kalman gain to elementwise operations. We also exploit these approximations for the \gls{cru}. The following summarizes the approximations by \citet{becker2019recurrent} and the resulting simplified update equations. We refer to \citet{becker2019recurrent} for detailed derivations.

\paragraph{Observation Model} The dimension of the latent state $M$ is twice of the dimension of the latent observation space $D$, i.e. $M = 2D$. The observation model links both spaces and is fixed at $\textbf{H} = [\mathbf{I}_D, \mathbf{0}_D]$ where $\mathbf{I}_D$ denotes the $D$-dimensional identity matrix and $\mathbf{0}_D$ the $D$-dimensional zero matrix. The idea is to split the latent state into an \textit{observed part}, which extracts information directly from the observations and a \textit{memory part}, which encodes features inferred over time. 

\paragraph{State Covariance} The covariance matrix of the latent state $\boldsymbol{\Sigma}_t$ is built of diagonal blocks $\boldsymbol{\Sigma}_t^{u}, \boldsymbol{\Sigma}_t^{l}, \boldsymbol{\Sigma}_t^{s}$, whose vectors of diagonal entries are denoted by $\boldsymbol{\sigma}_t^{u}, \boldsymbol{\sigma}_t^{l}, \boldsymbol{\sigma}_t^{s}$, respectively. 

\begin{equation}
    \boldsymbol{\Sigma}_t =
    \begin{pmatrix}
    \boldsymbol{\Sigma}_t^{u}  & \boldsymbol{\Sigma}_t^{s} \\
    \boldsymbol{\Sigma}_t^{s}  & \boldsymbol{\Sigma}_t^{l}
    \end{pmatrix}
\end{equation}

The observation part of the latent state is thus only correlated with the corresponding memory part. The argument behind this strong assumption is that the free parameters in the neural encoder and decoder suffice to find a representation where the above limitations hold. 

\paragraph{Simplified Update Step} The Kalman gain simplifies to a structure with two diagonal blocks of size $D \times D$, i.e. $\textbf{K}_t = \begin{bmatrix}
\textbf{K}_t^{u} & \textbf{K}_t^{l} 
\end{bmatrix}^{T}$. The vector of diagonal entries $\textbf{k}_t^{u}$, $\textbf{k}_t^{l}$ can be computed with element wise division ($\oslash$)

\begin{equation}
    \textbf{k}_t^{u} = \boldsymbol{\sigma}^{u,-}_{t} \oslash  (\boldsymbol{\sigma}_t^{u,-} + \boldsymbol{\sigma}_t^{\mathrm{obs}}) \qquad
    \textbf{k}_t^{l} = \boldsymbol{\sigma}^{s,-}_{t} \oslash  (\boldsymbol{\sigma}_t^{u,-} + \boldsymbol{\sigma}_t^{\mathrm{obs}}) 
\end{equation}

The posterior mean update then simplifies to 

\begin{equation}
    \boldsymbol{\mu}^+_t = \boldsymbol{\mu}^-_t + 
    \begin{bmatrix}
    \textbf{k}^{u}_{t} \\ 
    \textbf{k}^{l}_{t}
    \end{bmatrix}
    \odot
    \begin{bmatrix}
    \textbf{y}_t - \boldsymbol{\mu}^{u,-}_{t} \\ 
    \textbf{y}_t - \boldsymbol{\mu}^{u,-}_t
    \end{bmatrix}  
\end{equation}

where $\boldsymbol{\mu}_t^{u}$ and $\boldsymbol{\mu}_t^{l}$ denote the upper and lower part of the prior mean respectively, i.e. $\boldsymbol{\mu}_t^{-} =  \begin{bmatrix}
\boldsymbol{\mu}_t^{u,-} & \boldsymbol{\mu}_t^{l,-}
\end{bmatrix}^{T}$. The update of the posterior covariance reduces to

\begin{equation}
    \boldsymbol{\sigma}_t^{u,+} = (\textbf{1}_{m} - \textbf{k}_t^{u}) \odot \boldsymbol{\sigma}^{u,-}_t
\end{equation}
\begin{equation}
    \boldsymbol{\sigma}_t^{s,+} = (\textbf{1}_{m} - \textbf{k}_t^{u}) \odot \boldsymbol{\sigma}^{s,-}_t
\end{equation}
\begin{equation}
    \boldsymbol{\sigma}_t^{l,+} = \boldsymbol{\sigma}_{t}^{l,-} - \textbf{k}_t^{l} \odot \boldsymbol{\sigma}^{s,-}_t
\end{equation}

where $\odot$ denotes elementwise multiplication. 

\section{Log-likelihood Results}
\label{sec:loglikelihood}
To assess uncertainty computation in the presence of high observation noise, we report negative log-likelihood on pendulum regression.  \Cref{loglikelihood} shows Gaussian \gls{nll} on test data for methods providing uncertainty estimates. As for \gls{mse} results, \gls{cru} outperforms baseline models. 
\begin{table}[ht!]
%\footnotesize
\caption[Pendulum interpolation results]{Test Gaussian \gls{nll} (mean $\pm$ std) on pendulum regression.}
\begin{center}

\begin{tabular}{lcc}
Model &  Regression   \\
\toprule
\gls{gru}  & -4.78 $\pm$ 0.48\\
\gls{gru}-$\Delta_t$  &  -5.45 $\pm$ 0.09\\
\gls{rkn} &  -4.38 $\pm$ 0.82\\
\gls{rkn}-$\Delta_t$  & -5.26 $\pm$ 0.24\\
%GRU-ODE-B & \\
\gls{fcru} & 
-5.46 $\pm$	0.11
\\
\gls{cru} &  \textbf{-5.49 $\pm$ 0.05}\\
\toprule \\
\label{loglikelihood}
\end{tabular}
\end{center}
\vskip -.6in
\end{table}

\section{Implementation Details}
\label{implementation_details}
\paragraph{Training} In all experiments, we train each model for 100 epochs using the Adam optimizer \citep{kingma2014adam}. Reported MSE and Gaussian \gls{nll} results are averages over 5 runs. We used a batch size of 50 for the pendulum and \gls{ushcn} data and a batch size of 100 for Physionet. For \gls{ushcn} and Physionet, we split the data into 80\% train and 20\% test and used 25\% of the train set for validation. For the pendulum experiments, we generated 2 000 training sequences, 1 000 validation sequences and report results on a hold-out set of 1 000 sequences. The folds are reused for each compared model.

\paragraph{Hyperparameters and Architecture} Here, we summarize hyperparameter choices made in the empirical study with details provided in \Cref{implementation_grud,implementation_latentode,implementation_odernn,implementation_gruodeb,implementation_rkn,implementation_gru,implementation_cru,implementation_mtan}. We used the following procedure to select latent state size, number of layers, number of hidden units, and training parameters: 
For \gls{gru}-D, latent \gls{ode} and \gls{ode}-\gls{rnn}, we use the choices optimized by \citet{rubanova2019latent} in their Physionet setup. For \gls{mtand}, we keep the hyperparameters chosen by the authors on Physionet and adjust the number of hidden units to control for model size. We then designed the encoder and decoder of the \gls{cru} such that the \gls{cru} has roughly the same number of parameters as the latent ODE model. We proceed analogously for hyperparameters of GRU-ODE-B: We employ the hyperparameter settings optimized by the authors on their \gls{ushcn} experiment. We keep the baseline architectures fixed across experiments, except for the latent state size, which we chose separately for each experiment according to the dimensionality of the input and previous work. See \Cref{implementation_cru} for the latent state size of each experiment, which is shared across architectures. For the pendulum image set, encoder, decoder and latent space sizes of \gls{cru} follow the \gls{rkn} \citep{becker2019recurrent} baseline on this set. To scale the methods based on \glspl{node} to images, we also embedded ODE-RNN, latent ODE and GRU-ODE-B into the same encoder-decoder structure as \gls{cru}. Similarly, the GRU baseline is augmented with the same encoder and decoder. This was also done in the pendulum experiments of \citet{becker2019recurrent} and more details and justification can be found there. The transformer method, \gls{mtand}, scales to high-dimensional input and we thus apply it directly on the raw images. We found GRU-ODE-B unstable when trained jointly with an encoder and decoder on the image interpolation task and therefore trained it directly on the raw images. 

%\paragraph{Hyperparameters} For GRU-D, latent ODE and ODE-RNN, we used the training hyperparameters selected by \citet{rubanova2019latent} in their Physionet experiment. We reuse the hyperparameter setting on all datasets.
\subsection{mTAND}
\label{implementation_mtan}
We use a hidden state size of $M=30$ for pendulum, $M=20$ for Physionet, and $M=10$ for \gls{ushcn}. We use a learning rate of 0.001 and 64 reference time points as in \citet{ShuklaM21}. For interpolation and extrapolation, we train \gls{mtand}-Full with 10 hidden units in encoder and decoder resulting in a model that has still three times as many parameters as \gls{cru}. For the per-time-point regression task, we use \gls{mtand}-Enc with 10 hidden units. 

\subsection{RKN}
\label{implementation_rkn}
We use a hidden state size of $M=30$ for pendulum, $M=20$ for Physionet, and $M=10$ for \gls{ushcn}. For the pendulum experiment, we keep the architecture and hyperparameter choices from \citet{becker2019recurrent}. We employ the same choices for \gls{cru} across all experiments, which are detailed in \Cref{implementation_cru}. For the time-aware variant, \gls{rkn}-$\Delta_t$, we feed the time gaps as additional input to the transition net that learns weights $\alpha_{\tau(t)}$ for the basis matrices (\Cref{eqn:parameterization1}), i.e. $\alpha_{\tau(t)} = w_\psi([t-\tau(t), \boldsymbol{\mu}_{\tau(t)}^{+}])$.

\subsection{GRU}
\label{implementation_gru}
To make parameter sizes comparable, we use a hidden state size of 75 for GRU and \gls{gru}-$\Delta_t$ on all datasets. To test GRU on image data, we embed a GRU cell in the encoder-decoder architecture employed for RKN and CRU as in \citet{becker2019recurrent}.  For \gls{gru}-$\Delta_t$, the time gap between observations is concatenated to the input of the \gls{gru} cell.

\subsection{GRU-D}
\label{implementation_grud}
We use a hidden state size of $M=30$ for pendulum, $M=20$ for Physionet, and $M=10$ for \gls{ushcn}. The other parameters are fixed across experiments. As described previously, we use hyperparameter and architecture choices by \citet{rubanova2019latent}, which result in 100 hidden units, a learning rate of 0.01 with a decay factor of 0.999.

\subsection{Latent ODE}
\label{implementation_latentode}
We use a latent state size of $M=30$ for pendulum, $M=20$ for Physionet, and $M=10$ for \gls{ushcn}. The other parameters are fixed across experiments. We used the latent ODE with an ODE-RNN recognition model. The recognition model has a hidden state of 40 dimensions, an ODE function with 3 layers and 50 hidden units and a GRU update with 50 hidden units. The ODE function of the generative model has 3 layers with 50 hidden units. We train the method with a learning rate of 0.01 and a decay rate of 0.999.

\subsection{ODE-RNN}
\label{implementation_odernn}
We use a hidden state size of $M=30$ for pendulum, $M=20$ for Physionet and $M=10$ for \gls{ushcn}. The other parameters are fixed across experiments. We use the hyperparameter choices proposed by \citet{rubanova2019latent}. Notably, we use 100 hidden units and a learning rate of 0.01 with a decay rate of 0.999. 

\subsection{GRU-ODE-B}
\label{implementation_gruodeb}
We use a latent state size of $M=30$ for pendulum, $M=20$ for Physionet and $M=10$ for \gls{ushcn}. The other parameters are fixed across experiments. As outlined above, we keep the hyperparameters determined by the authors in their \gls{ushcn} experiment. That is to say, we use 50 hidden units, a learning rate of 0.001, weight decay of 0.0001, and a dropout rate of 0.2. The $f_{prep}$ function of the GRU-Bayes component has 10 hidden units, the $f_{obs}$ mapping has 25 hidden units.

\subsection{CRU and \gls{fcru}}
\label{implementation_cru}

We trained \gls{cru} with an Adam optimizer \citep{kingma2014adam} on the Gaussian negative log-likelihood (\Cref{eqn:rkn4}) for Physionet, \gls{ushcn} and pendulum angle prediction and on the loss of  \Cref{eqn:rkn5} for the pendulum interpolation task. On the validation set in the Pendulum interpolation experiment, we found a learning rate of 0.001 to work best for \gls{cru} and a slightly higher learning rate of 0.005  for \gls{fcru}. We kept this choice for all datasets throughout all other experiments. (Through hyperparameter optimization on each dataset separately, the experimental results might improve further.) Gradient clipping was used on \gls{ushcn} and Physionet. 

The initial conditions for the latent state are set to $\boldsymbol{\mu}^{-}_{t_0} = \mathbf{0}$ and $\mathbf{\Sigma}^{-}_{t_0} = 10 \cdot \mathbf{I}$. We found an initialization of the transitions such that the prior mean is close to the posterior mean of the previous time step crucial for performance and stability. Thus, we initialized the basis matrices $\{\mathbf{A}^{(k)}\}_{k=1...K}$ filled with zeros to start off with a prediction step of $\boldsymbol{\mu}^-_{t'} = \mathbf{I} \boldsymbol{\mu}^+_t$. Equivalently, we chose $\mathbf{E} = \mathbf{I}$ and $\mathbf{D}^{(k)} = 1e^{-5} \cdot \mathbf{I}, \forall k=1...K$ for the \gls{fcru} initialization. Missing features are zero-encoded and masked out in the \gls{nll} and \gls{mse} computation. For all experiments, we used a transition net $w_\psi$ with one linear layer and softmax output. 

The \gls{cru} architecture used in each experiment is explicitly summarized next.

\subsubsection{Pendulum Interpolation}

\paragraph{Continuous-discrete Kalman filter}
\begin{itemize}
\vspace{-5pt}
\setlength\itemsep{0em}
    \item Latent observation dimension: 15
    \item Latent state dimension: 30
    \item Number of basis matrices: 15
    \item Bandwidth (for \gls{cru}): 3
\end{itemize}

\paragraph{Encoder:} 2 convolution, 1 fully connected, linear output
\begin{itemize}
\setlength\itemsep{0em}
    \item  Convolution, 12 channels, 5 $\times$ 5 kernel, padding 2, ReLU, max pooling with 2 $\times$ 2 kernel and 2 $\times$ 2 stride
    \item  Convolution, 12 channels, 3 $\times$ 3 kernel, padding 1, 2 $\times$ 2 stride, ReLU, max pooling with 2 $\times$ 2 kernel and 2 $\times$ 2 stride
    \item  Fully-connected, 30 neurons, ReLU 
    \item  Linear output for latent observation; linear output, elu+1 activation for latent observation variance
\end{itemize}

\paragraph{Decoder output sequence $\mathbf{o}_\mathcal{T}$:} 1 fully-connected, 3 Transposed convolution
\begin{itemize}
\setlength\itemsep{0em}
    \item  Fully connected, 144 neurons, ReLU
     \item  Transposed convolution, 16 channels, 5 $\times$ 5 kernel, padding 2, 4 $\times$ 4 stride, ReLU
     \item  Transposed convolution, 12 channels, 3 $\times$ 3 kernel, padding 1, 2 $\times$ 2 stride, ReLU
     \item  Transposed convolution, 1 channel, 2 $\times$ 2 kernel, padding 5, 2 $\times$ 2 stride, sigmoid activation
\end{itemize}

\subsubsection{Pendulum Regression}
We used the same encoder and Kalman filter architecture as in the pendulum interpolation task.

\paragraph{Decoder output sequence $\mathbf{o}_\mathcal{T}$:} 1 fully-connected, linear output
\begin{itemize}
\setlength\itemsep{0em}
    \item  Fully connected, 30 neurons, Tanh
    \item  Linear output
\end{itemize}

\paragraph{Decoder output variance $\boldsymbol{\sigma}^{out}_\mathcal{T}$:} 1 fully-connected, linear output
\begin{itemize}
\setlength\itemsep{0em}
    \item  Fully connected, 30 neurons, Tanh
    \item  Linear output, elu+1 activation
\end{itemize}

\subsubsection{USHCN}
\paragraph{Continuous-discrete Kalman filter}
\begin{itemize}
\setlength\itemsep{0em}
    \item Latent observation dimension: 5
    \item Latent state dimension: 10
    \item Number of basis matrices: 15
    \item Bandwidth (for \gls{cru}): 3
\end{itemize} 

\paragraph{Encoder:} 3 fully connected, linear output
\begin{itemize}
\setlength\itemsep{0em}
    \item  Fully connected, 50 neurons, ReLU, layer normalization
    \item  Fully connected, 50 neurons, ReLU, layer normalization
    \item  Fully connected, 50 neurons, ReLU, layer normalization
    \item Linear output for latent observation; linear output, square activation for latent observation variance
\end{itemize}

\paragraph{Decoder output sequence $\mathbf{o}_\mathcal{T}$:} 3 fully connected, linear output
\begin{itemize}
\setlength\itemsep{0em}
    \item  Fully connected, 50 neurons, ReLU, layer normalization
    \item  Fully connected, 50 neurons, ReLU, layer normalization
    \item  Fully connected, 50 neurons, ReLU, layer normalization
    \item  Linear output 
\end{itemize}

\paragraph{Decoder output variance $\boldsymbol{\sigma}^{out}_\mathcal{T}$:} 1 fully-connected, linear output
\begin{itemize}
\setlength\itemsep{0em}
    \item  Fully connected, 50 neurons, ReLU, layer normalization
     \item  Linear output, square activation
\end{itemize}

\subsubsection{Physionet}
We used the same encoder and decoder architecture as in the \gls{ushcn} experiment.
\paragraph{Continuous-discrete Kalman filter}
\begin{itemize}
\setlength\itemsep{0em}
    \item Latent observation dimension: 10
    \item Latent state dimension: 20
    \item Number of basis matrices: 20
    \item Bandwidth (for \gls{cru}): 10
\end{itemize}

\section{Data Preprocessing}
\subsection{USHCN} Daily weather records can be downloaded at \url{https://cdiac.ess-dive.lbl.gov/ftp/ushcn_daily/}. We remove observations with a bad quality flag as in \citet{brouwer2019gru}. However, unlike \citet{brouwer2019gru} we are interested in long term extrapolation and thus, select a different time window of four years from 1990 to 1993. We keep only centers that start reporting before 1990 and end reporting after 1993. We split the remaining 1168 centers into 60\% train 20\% validation and 20\% test set. For each set, we remove measurements that are more than four standard deviations away from the set mean and normalize each feature to be in the [0,1] interval individually per set. For the baselines, we apply the time scaling strategies of previous work: we scale timestamps to be in the [0,1] for ODE-RNN, latent ODE and GRU-D as in \citet{rubanova2019latent} and feed time points unprocessed to GRU-ODE-B as in \citet{brouwer2019gru}. For \gls{cru}, we scale the timestamps, which unit is days, by a factor of 0.3.

\subsection{Physionet} The data is publicly available for download at \url{https://physionet.org/content/challenge-2012/1.0.0/}. We preprocess the data as in  \citet{rubanova2019latent}: We discard four general descriptors reported once at admission (age, gender, height, ICU-type) and keep only the remaining set of 37 time-variant features. Time points are rounded to 6 minutes steps. We split the patients into 60\% train, 20\% validation and 20\% test set. Lastly, we normalize each feature to be in the [0,1] interval separately per set. To mimic the training routine proposed by the authors, we scale the timestamps to the [0,1] interval for GRU-D, ODE-RNN and latent ODE, leave the timescale unchanged for GRU-ODE-B, and multiply timestamps by 0.2 for \gls{cru}.

\section{Computing Infrastructure}
Models were trained on one Nivida TU102GL Quatro RTX 6000/8000 with 40 physical Intel Xeon Gold 6242R CPU. 

\section{Source Code}
Our code is available at \url{https://github.com/boschresearch/Continuous-Recurrent-Units}. 
\section{Sample Trajectory}
\begin{figure}[h!]
  \begin{subfigure}{}
    \includegraphics[width=\linewidth]{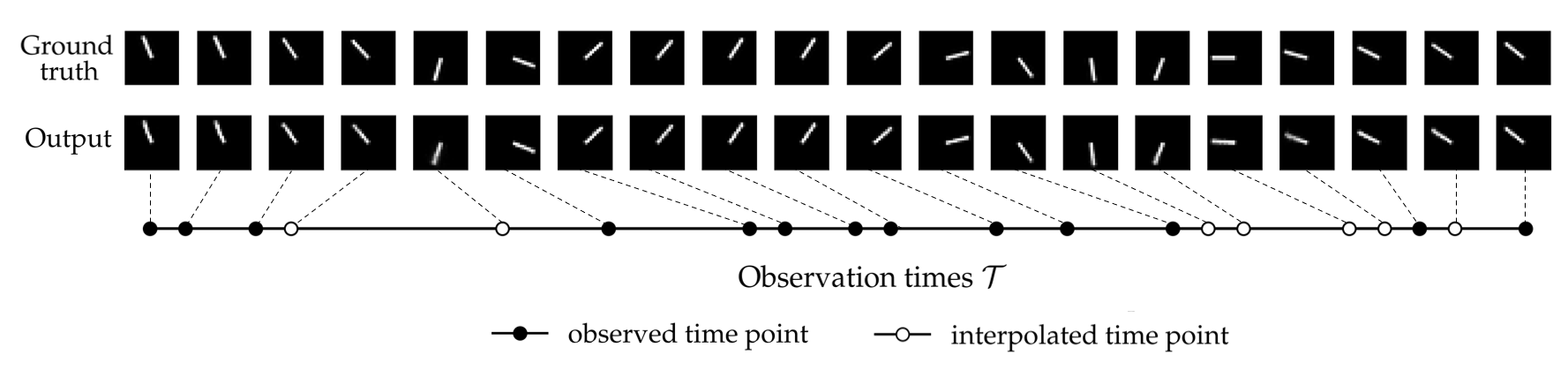}
    \caption{\gls{fcru}} \label{fig:1b}
  \end{subfigure}%
\caption{Test trajectory for the pendulum interpolation task: The \gls{fcru} predicts images precisely despite irregular intervals between image frames.}
\label{fig:pendulumii}
\end{figure}

\end{document}